\documentclass[sigplan,screen]{acmart}
  
\usepackage{booktabs} 

\usepackage{balance}

\setcopyright{none}
\acmDOI{}
\acmYear{2018}
\copyrightyear{2018}
\acmISBN{--}
\acmConference{Intel Labs}{MIT}{2018}

\renewcommand\footnotetextcopyrightpermission[1]{} 

\begin{document}
\title{The Three Pillars of Machine Programming}

\author{Justin Gottschlich}
\affiliation{%
  \institution{Intel Labs, USA}
}
\email{justin.gottschlich@intel.com}

\author{Armando Solar-Lezama}
\affiliation{%
  \institution{MIT, USA}
}
\email{asolar@csail.mit.edu}

\author{Nesime Tatbul}
\affiliation{%
  \institution{Intel Labs and MIT, USA}
}
\email{tatbul@csail.mit.edu}

\author{Michael Carbin}
\affiliation{%
  \institution{MIT, USA}
}
\email{mcarbin@csail.mit.edu}

\author{Martin Rinard}
\affiliation{%
  \institution{MIT, USA}
}
\email{rinard@csail.mit.edu}

\author{Regina Barzilay}
\affiliation{%
  \institution{MIT, USA}
}
\email{regina@csail.mit.edu}

\author{Saman Amarasinghe}
\affiliation{%
  \institution{MIT, USA}
}
\email{saman@csail.mit.edu}

\author{Joshua B. Tenenbaum}
\affiliation{%
  \institution{MIT, USA}
}
\email{jbt@mit.edu}

\author{Tim Mattson}
\affiliation{%
  \institution{Intel Labs, USA}
}
\email{timothy.g.mattson@intel.com}

\renewcommand{\shortauthors}{Gottschlich et al.}

\begin{abstract}
In this position paper, we describe our vision of the future of machine programming through a categorical examination of three pillars of research. Those pillars are: \emph{(i)} intention, \emph{(ii)} invention, and \emph{(iii)} adaptation. Intention emphasizes advancements in the human-to-computer and computer-to-machine-learning interfaces. Invention emphasizes the creation or refinement of algorithms or core hardware and software building blocks through machine learning (ML). Adaptation emphasizes advances in the use of ML-based constructs to autonomously evolve software.
\end{abstract}

\keywords{program synthesis, machine programming, software development, software maintenance, intention, invention, adaptation}

\maketitle

\section{Introduction}

Programming  is a cognitively demanding task that requires extensive knowledge, experience and a large degree of creativity, and is notoriously difficult to automate. Machine learning (ML) has the capacity to reshape the way software is developed. At some level, this has already begun, as machine-learned components progressively replace complex hand-crafted algorithms in domains such as natural-language understanding and vision. Yet, we believe that it is possible to move much further. We envision machine learning and automated reasoning techniques that will enable new programming systems; systems that will deliver a significant degree of automation to reduce the cost of producing secure, correct, and efficient software. These systems will also enable non-programmers to harness the full power of modern computing platforms to solve complex problems correctly and efficiently. We call such efforts \emph{machine programming}.

\begin{figure}[htpb]
\begin{center}
\includegraphics[width=1.0\columnwidth]{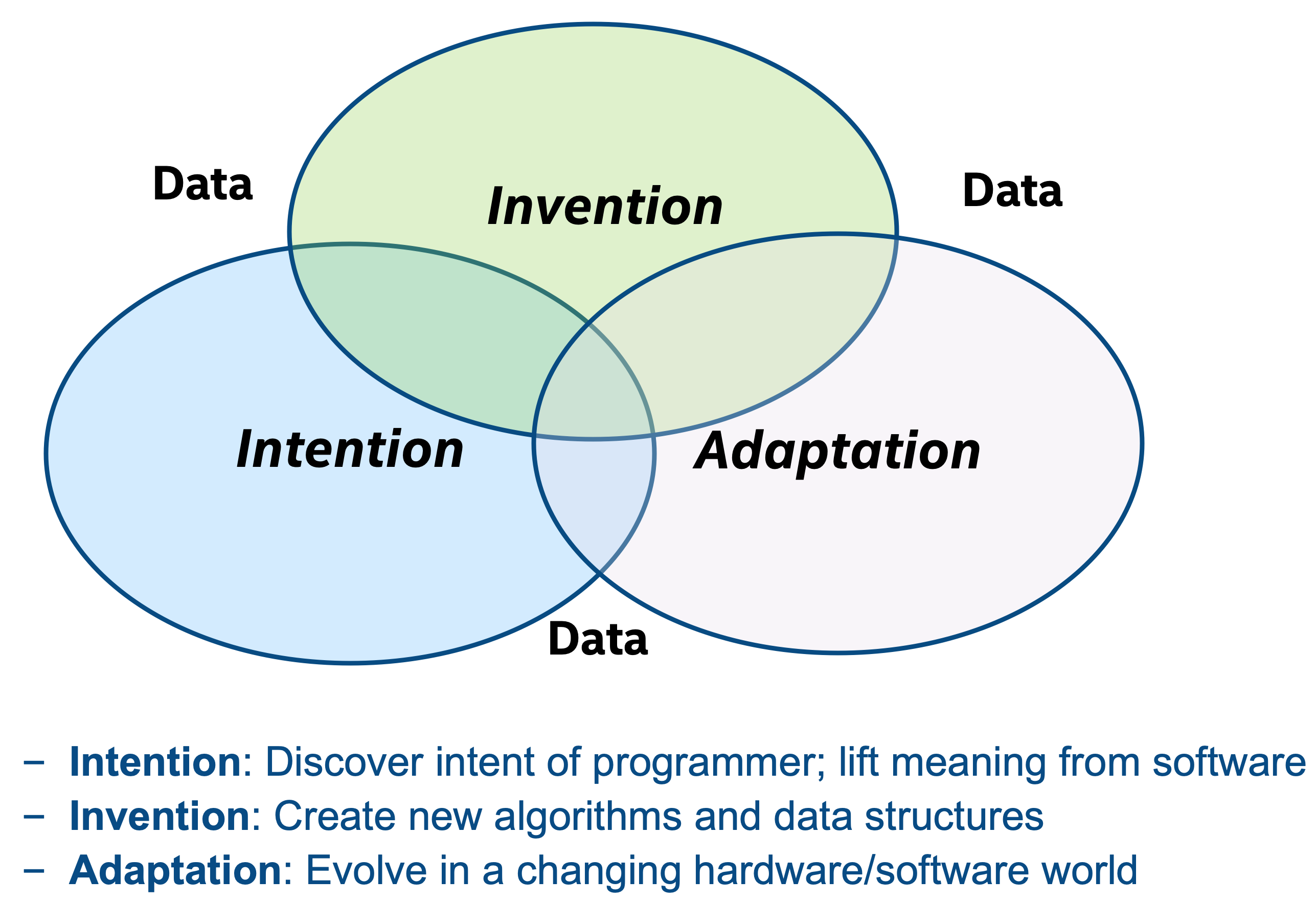}
\vspace{-0.2cm}
\caption{The Three Pillars of Machine Programming: Intention, Invention, and Adaptation. Each pillar in the diagram includes a few example sub-domains generally related to them.}
\label{fig:pillars}
\end{center}
\end{figure}

\subsection{Why Now?}

Programming is the process of turning a problem definition (the \emph{intent}) into a sequence of instructions that when executed on a computer, produces a solution to the original problem. Over time, a program must be \emph{maintained} as it adapts to changes in the program's goals, errors in the program, and the features in new computer platforms. A \emph{machine programming system} is any system that automates some or all of the steps of turning the user's intent into an executable program and maintaining that program over time.


The automation of programming has been a goal of the programming systems community since the birth of Fortran in the 1950s. The first paper on \emph{``The FORTRAN Automatic Coding System''} made it clear that its goal was ``for the 704 [IBM's next large computer] to code problems for itself and produce as good programs as human coders (but without the errors)''~\cite{Backus57}. The broader AI community has also been interested in automatic programming dating back to the Programmer's Apprentice Project back in the late 1970s~\cite{rich:1988:par}. A number of technological developments over the past few years, however, are creating both the \emph{need} and the \emph{opportunity} for transformative advances in our ability to use machines to help users write software programs.

\paragraph{Opportunity}
Humans interact through speech, images, and gestures; so-called ``natural inputs''.
Advances in deep learning and related machine learning technologies have dramatically improved a computer's ability to associate meaning with natural inputs.  Deep learning also 
makes it possible to efficiently represent complex distributions over classes of structured objects; a crucial capability if one wants to automatically synthesize a program using probabilistic or direct transformation techniques. In parallel to advances in machine learning, the programming systems community has been making notable advances in its ability to reason about programs and manipulate them. Analyzing thousands of lines of code to derive inputs that expose a bug has moved from an intractable problem into one that is routinely solved due to advances in automated reasoning tools such as SAT and SMT solvers. Data, a key enabler for learning-based strategies, is also more available now than at any time in the past. This is the byproduct of at least two factors: \emph{(i)} the emergence of code repositories, such as GitHub, and \emph{(ii)} the growing magnitude of the web itself, where it is possible to observe and analyze the code (e.g., JavaScript) powering many web applications. Finally, the advent of cloud computing makes it possible to harness large-scale computational resources to solve complex analysis and inference problems that were out of reach only a few years ago.

\paragraph{Need} The end of Dennard scaling means that performance improvements now come through increases in the complexity of the hardware, with resulting increases in the complexity of  compilation targets~\cite{esmae:2011:dennard}.
Traditional compilation techniques rely on an accurate model of relatively simple hardware. These techniques are inadequate for exploiting the full potential of heterogeneous hardware platforms. The time is ripe for techniques, based on modern machine learning, that learn to map computations onto multiple platforms for a single application. Such techniques hold the promise of effectively working in the presence of the multiple sources of uncertainty that complicate the use of traditional compiler approaches. 
Moreover, there is a growing need for people with core expertise outside of computer science to program, whether for the purpose of data collection and analysis, or just to gain some control over the growing set of digital devices permeating daily life. 

\subsection{The Three Pillars}
Given the opportunity and the need, there are already a number of research efforts in the direction of machine programming in both industry and academia.
The general goal of machine programming is to remove the burden of writing correct and efficient code from a human programmer and to instead place it on a machine. The goal of this paper is to provide a conceptual framework for us to reason about machine programming. We describe this framework in the context of three technical pillars: \emph{(i)} intention, \emph{(ii)} invention, and \emph{(iii)} adaptation. Each of these pillars corresponds to a class of capabilities that we believe are necessary to transform the programming landscape. 

\emph{Intention} is the ability of the machine to understand the programmer's goals through more natural forms of interaction, which is critical to reducing the complexity of writing software. \emph{Invention} is the ability of the machine to discover how to accomplish such goals; whether by devising new algorithms, or even by devising new abstractions from which such algorithms can be built. \emph{Adaptation} is the ability to autonomously evolve software, whether to make it execute efficiently on new or existing platforms, or to fix errors and address vulnerabilities. As suggested in Figure~\ref{fig:pillars}, intention, invention, and adaptation intersect in interesting ways. When advances are made in one pillar, another pillar may be directly or indirectly influenced. In Section~\ref{sect:interplay} we show that sometimes such influence can be negative. This further emphasizes the importance for the machine programming research community to be cognizant of these pillars moving forward and to understand how their research interacts with them.

The remainder of this report is organized as follows. In Sections~\ref{sect:intention}, \ref{sect:invention}, \ref{sect:adaptation}, we provide a detailed examination of intention, invention, and adaptation, respectively. We also discuss the interactions between them, throughout. In Section~\ref{sect:interplay}, we provide a concrete analysis of verified lifting~\cite{kamil:2016:VLS} and how it interacts with each of the three pillars (in some cases, disruptively). We close with a discussion on the impact of data, as it is the cornerstone for many ML-based advances.

\section{Intention}
\label{sect:intention}

\emph{Intention} corresponds to the class of challenges involved in capturing the user's intent in a way that does not require substantial programming expertise. One of the major challenges in automating programming is capturing the user's intent; describing any complex functionality to the level of detail required by a machine quickly becomes just programming by another name. Table~\ref{tab:intention} provides a brief overview of existing research in the space of intention. It consists of three columns: \emph{Research Area}, \emph{System}, and \emph{Influence}. The \emph{Research Area} column includes subdomains of research for the given pillar. The \emph{System} column includes a non-exhaustive list of examples of systems for that subdomain. The \emph{Influence} column lists the different pillars, other than intention, that are influenced by the system listed in the corresponding \emph{System} column.~\footnote{This table is not meant as an exhaustive survey of all the research in the intention pillar. Rather, it is meant as an example of work in subdomains of intention.}
This table structure is also used for the invention and adaptation pillar sections.

It is useful to contrast programming with human-human interactions, where we are often able to convey precise intent by relying on a large body of shared context. If we ask a human to perform a complex, nuanced task such as generating a list of researchers in the ML domain, we can expect the hidden details not provided in our original description to be implicitly understood (e.g., searching the internet for individuals both in academia and industry with projects, publications, etc., in the space of ML). By contrast, writing a computer program to do this would be a more significant undertaking because we would have to explicitly detail how to accomplish each step of the process. Libraries provide some assistance, but libraries themselves are difficult to write, can be difficult to use, and, in many cases, even difficult to find.

\begin{table}[h]
\centering
\caption{Examples of Research in the Intention Pillar.}
\label{tab:intention}
\begin{tabular}{lccc}
\toprule
\textbf{Research Area} & \textbf{System} & \textbf{Influence} \\
\midrule
Examples & Domain-Specific &  \\
 & Input-Output ML~\cite{chen:2017:example} & -- \\
 & FlashFill~\cite{gulwani:2011:flashfill} & Invention \\
\midrule
Generalizability  & Recursion~\cite{CaiSS17} & Adaptation \\
\midrule
Natural Language & Babble Labble~\cite{babble:re:2017} & -- \\
& SQLNet~\cite{xu:2017:natural} & -- \\
& NL2P~\cite{NL2P} & -- \\
\midrule
Partial & Sketch~\cite{Solar-LezamaTBSS06} & Adaptation \\
Implementations & AI Programmer~\cite{becker:2017:ai} & Adaptation \\
\bottomrule
\end{tabular}
\end{table}

For the last decade, the program synthesis community has struggled with this problem, both in the broad context of synthesizing programs from specifications, as well as in the narrower context of inductive programming or programming by example~\cite{muggleton:1991:ip, cypher:1993:wwid}. There are at least two major observations that have emerged from prior work. The first observation is that by tightly controlling the set of primitives from which programs can be built and imposing strong biases on how these primitives should be connected, it is possible to cope with significant ambiguities in the specification. For example, the work on FlashFill~\cite{gulwani:2011:flashfill} demonstrated that it is possible to synthesize a desired string manipulation from a small number of examples, often only one, by restricting the space of programs to a carefully crafted domain specific language (DSL) with carefully tuned biases. Moreover, subsequent work showed that such biases could be learned, rather than having to be tailored by hand~\cite{EllisG17,SinghG15}. Similar observations have been made in other contexts, from the synthesis of SQL queries, to the synthesis of Java APIs~\cite{Yaghmazadeh0DD17,vijay:2017:bayou}. The Bayou project~\cite{vijay:2017:bayou}, for example, shows that it is possible to use deep neural networks to learn complex conditional distributions over programs, allowing a user to generate complex Java code from concise traces of evidence. Models not based on neural networks can also be used to learn distributions over programs for this purpose~\cite{BielikRV16}.

The second major observation is that multi-modal specification can help in unambiguously describing complex functionality. One of the earliest examples of multi-modal synthesis was the Storyboard Programming Tool (SPT), which allowed a user to specify complex data-structure manipulations through a combination of abstract diagrams, concrete examples, and code skeletons~\cite{SinghS11,SinghS12}. The observation was that fully specifying a transformation via any of these modalities on its own, code, examples or abstract diagrams, was difficult, but in combination each of these formalisms could cover for the shortcomings of the other. A similar observation was made by the Transit project, which showed that it was possible to synthesize complex cache coherence protocols from a combination of temporal properties, concrete and symbolic examples~\cite{UdupaRDMMA13}. 

Addressing the intention challenges more fully, however, will require additional breakthroughs at the intersection of machine learning and programming systems. One of the major opportunities is exploiting the ability modern learning techniques to extract meaning from high-dimensional unstructured inputs, such as images or speech. Recent work, for example on converting natural-language to programs, has shown the potential for exploiting this in the context of narrow domains~\cite{Yaghmazadeh0DD17,NL2P,IyerKCKZ17}. Similarly, recent work on extracting programmatic representations from hand-drawn images has demonstrated the possibilities of using visual data as a basis for conveying intent~\cite{EllisRST17}. Many of these systems, however, are one-off efforts targeted at narrow domains; one of the major questions is how to support this kind of functionality while maintaining the versatility of modern programming systems, and how to scale such high-level interactions to richer more complex tasks, including tasks that may require input from more than one person to fully describe.

\section{Invention}
\label{sect:invention}

Invention emphasizes the creation or refinement of algorithms or core hardware and software building blocks. For program construction, invention usually involves generating the series of steps that a machine would have to execute to fulfill a user's intent; in essence, it is the process of generating algorithms. This may require discovering new algorithms that are unique and different from prior contributions within the same space. In many instances, however, invention will be accomplished by identifying how to combine and adapt known data structures and algorithmic primitives to solve a particular problem. Both the program synthesis and the machine learning communities have made notable progress in this space in recent years, but there remain many open problems to be solved.  See Table~\ref{tab:invention} for highlights of existing research in the space of invention.~\footnote{This table is not meant as an exhaustive survey of all the research in the invention pillar. Rather, it is meant as an example of work in subdomains of invention.}

\begin{table}[h]
\centering
\caption{Examples of Research in the Invention Pillar.}
\label{tab:invention}
\begin{tabular}{lccc}
\toprule
\textbf{Research Area} & \textbf{System} & \textbf{Influence} \\
\midrule
Explicit Search &   $\lambda^2$~\cite{FeserCD15}    & Intention \\
      & SynQuid~\cite{PolikarpovaKS16}    & Intention  \\
\midrule
Constraint-Based  & Sketch~\cite{Solar-LezamaTBSS06} & Intention \\
        ~         & PTS~\cite{SrivastavaGF10}                & Intention \\
 \midrule
Symbolic Version & & \\
Space \hspace{-12pt} & FlashFill~\cite{gulwani:2011:flashfill} & Intention \\
\midrule
Deductive & Paraglide~\cite{VechevY08}      & Adaptation \\
                  & Fiat~\cite{DelawarePGC15}       & Adaptation \\
                  & Spiral~\cite{PuschelMSXJPVJ04} & Adaptation \\
 \midrule
Learning Directed & DeepCoder~\cite{balog:2017:iclr} & Intention \\
& Bayou~\cite{vijay:2017:bayou}            & Intention \\
 \midrule
Learning to Learn & Learning to & \\
& Optimize~\cite{li:2017:iclr} & Adaptation \\
\bottomrule
\end{tabular}
\end{table}

\subsection{Program Synthesis}

For program synthesis, the modern approach has been to frame invention as a search problem where, given a space of candidate programs, the goal is to search for one that satisfies a set of constraints on the desired behavior~\cite{sygus15}. This type of research has focused on questions of \emph{(i)} how to represent the search space, \emph{(ii)} how to explore it efficiently by exploiting knowledge of the semantics of the underlying building blocks, as well as \emph{(iii)} understanding and advancing the structure of the semantic constraints themselves. 

At a high-level, researchers have explored at least two major classes of approaches to this problem. The first class involves search techniques that explicitly try to build a syntactic representation of each program in the search space---abstract syntax trees (ASTs) are common as a representation. These techniques achieve efficiency by ruling out large sets of possible programs without exploring them one-by-one, usually by discovering that particular sub-structures can never be part of a correct solution~\cite{PolikarpovaKS16,OseraZ15,UdupaRDMMA13,AlbarghouthiGK13,FeserCD15}. The second class involves symbolic search techniques, where the entire program space is represented symbolically, either using a special purpose representation~\cite{gulwani:2011:flashfill,PolozovG15}, or, in the case of \emph{constraint-based synthesis}, by reducing it to a set of constraints whose solution can be mapped to a concrete program, which can be solved using a SAT or SMT solver~\cite{Solar-LezamaTBSS06,SrivastavaGF10,GulwaniJTV11}, or in some cases a numerical optimization procedure~\cite{ChaudhuriS10}.

Many of these techniques, especially those designed to support rich specifications instead of simply input-output examples, have been enabled by the ability to automatically reason about the correctness of candidate programs, and in some cases, the ability to use static analysis to rule out large sets of candidate programs all at once~\cite{PolikarpovaKS16,OseraZ15}. 

These techniques have made tremendous progress in recent years; for example, in the domain of bit-level manipulation, the most recent winner of the Syntax Guided Synthesis competition (SyGuS Comp) was able to automatically discover complex bit-level manipulation routines that were considered intractable only a few years earlier~\cite{AlurCR15}. In the case of string manipulations, program synthesis is now robust enough to ship as part of commercial products (e.g. Flashfill in Excel~\cite{gulwani:2011:flashfill}).
In the context of data-structure manipulations, routines such as red-black tree insertion and complex manipulations of linked lists can now be synthesized and verified in the context of both imperative and functional languages~\cite{PolikarpovaKS16}, and in the functional programming realm, routines that were once considered functional pearls can now be synthesized from a few examples~\cite{FeserCD15}. 

That said, there are fundamental limitations to the recent program synthesis approach to invention. Even with a restrictive set of primitives, the search-space grows exponentially with the size of the code-fragments that one aims to discover, making it difficult to scale beyond a dozen or so lines of code. There are some instances of systems that have been able to discover more complex algorithms, either by building them incrementally~\cite{PerelmanGGP14}, or by breaking down the problem into smaller pieces---either by providing the synthesizer with the interfaces of sub-components to use~\cite{PolikarpovaKS16}, or by leveraging some domain-specific structure of the problem to decompose it into a large number of independent sub-problems~\cite{Inala0S16}. 

Deductive synthesis techniques are another class of approaches to the Invention problem, where the idea is to start with a high-level specification and refine it to a low-level implementation by applying deductive rules or semantics-preserving transformations. This class of techniques has proven to be successful, for example, in automating the development of concurrent data-structures~\cite{VechevY08} or signal processing pipelines~\cite{PuschelMSXJPVJ04}. The growing power of interactive theorem provers such as Coq have also made it possible to get strong correctness guarantees from code developed through this approach~\cite{DelawarePGC15}. The main drawback of this class of techniques is that while they tend not to suffer from the same scalability problems as the search-based techniques---because they break the problem into a number of small local reasoning steps---they tend to be domain specific, because they rely on carefully engineered deductive rules for the particular problem domain to operate effectively.

\subsection{Machine Learning}

Parallel to these efforts, the ML community has been exploring similar ideas in a different context. At one level, machine learning itself can be considered a form of invention. Many ML algorithms, including support vector machines (SVMs) and deep learning, can be seen as a form of constraint-based synthesis, where the space of programs is restricted to the set of parameters for a specific class of parameterized functions, and where numerical optimization is used to solve the constraints, which in this case involve minimizing an error term. By focusing on a narrow class of parameterized functions, machine-learning techniques are able to support search spaces that are larger than what the aforementioned synthesis techniques can support. Neural networks with a million real-valued parameters are becoming standard for many applications, whereas the largest problems solved by the SMT-based techniques have on the order of a thousand boolean parameters. This allows neural-networks to capture significantly more complexity. 

More recently, there have been significant efforts to capture more general program structure with neural networks, either by encoding differentiable Turing Machines~\cite{GravesWD14}, or by incorporating the notion of recursion directly into the neural network~\cite{CaiSS17}. However, when it comes to synthesizing programs with significant control structure and that require more discrete reasoning, the techniques from the synthesis community tend to outperform the neural network based techniques~\cite{GauntBSKKTT16}. 

\subsection{New Directions} \label{sect:invention-new}

A major opportunity for breakthroughs in the invention problem lies at the intersection of the two lines of research. One important idea that is beginning to emerge is the use of learning-based techniques to learn distributions over the space of programs that are conditioned on features from the stated goals of the desired program. These distributions can be used to narrow the space of programs to something tractable. For example, DeepCoder~\cite{balog:2017:iclr} uses a neural network to map from the intention (given as a set of examples) to a restricted set of components that it has learned to recognize as useful when satisfying similar intentions. This allows it to then use an off-the-shelf synthesizer to solve the synthesis problem on this restricted program space. The Bayou project uses a more sophisticated network architecture to learn much more complex conditional distributions, allowing it to automatically determine, for example, how to use complex Java and Android APIs~\cite{vijay:2017:bayou}. 

One of the open challenges in this space is to develop systems that can solve large-scale invention challenges, moving beyond simple algorithms and data-structure manipulations, by solving problems at the scale of an ACM programming competition or a collegiate programming course. This requires systems that can better mimic the way programmers approach these problems today. That is, using knowledge accumulated through practice and directed study to identify the core algorithmic building blocks needed to solve a problem. This also includes reasoning at a high-level of abstraction about how those building blocks fit together, and only then reasoning at the code level in a targeted fashion.

An important set of challenges in solving this problem is that while there are extensive resources to help humans learn to program, from tutorials to textbooks to stackoverflow.com, most of those resources are not suitable for data-hungry ML methods, such as deep learning. Applying ML in this domain may require a combination of new ML methods that can learn from data-sources aimed at humans, with novel solutions to exploit large-scale data sources, such as  code repositories like GitHub, or synthetic data-sources such as randomly generated programs and datasets. 

As architectures continue to evolve and become more complex and reconfigurable, some of the responsibility for coping with this complexity will fall on the invention layer, either because it will have to discover algorithms that map well to the constraints imposed by the hardware, or in the case of architectures that include FPGAs, the invention layer may need to derive the hardware abstractions themselves that align for a given algorithm. There is already some precedent in using constraint-based synthesis to handle non-standard architectures, ranging from exploiting vector instructions~\cite{BartheCKGM13}, to synthesizing for complex low-power architectures~\cite{PhothilimthanaJSTCB14}, but significantly more research is needed for this problem to be fully addressed.

\section{Adaptation}
\label{sect:adaptation}

Determining the algorithmic steps to solve a problem is only one part of the software development process. The resulting algorithms must be made to run efficiently on one or more target platforms,  and after the code is deployed in the field, the code must be maintained as users expose bugs or corner cases where the system does not behave as expected. Moreover, as workloads on the system evolve, it may be necessary to re-evaluate optimization decisions to keep the software running at its peak performance. Together, these capabilities make up the \emph{Adaptation} pillar. See Table~\ref{tab:adaptation} for highlights of existing research in the space of adaptation.~\footnote{This table is not meant as an exhaustive survey of all the research in the adaptation pillar. Rather, it is meant as an example of work in subdomains of adaptation.}

\begin{table}[h]
\centering
\caption{Examples of Research in the Adaptation Pillar.}
\label{tab:adaptation}
\begin{tabular}{lccc}
\toprule
\textbf{Research Area} & \textbf{System} & \textbf{Influence} \\
\midrule
Autotuning & OpenTuner~\cite{ansel:2014:opentuner} & -- \\
& PetaBricks~\cite{Ansel:2009:petabricks} & -- \\
\midrule
Code-to-Code & Verified Lifting~\cite{kamil:2016:VLS} & Intention \\
& Tree-to-Tree  &  \\
& Translation~\cite{chen:2017:tree} & Intention \\
\midrule
Correctness & ACT~\cite{alam:2016:act} & Intention \\
& CodePhage~\cite{CodePhage} & Intention \\
\midrule
Data Structures & Learned Index & \\
& Structures~\cite{kraska:2017:learned} & Invention \\
\midrule
Mathematics & SPIRAL~\cite{PuschelMSXJPVJ04} & -- \\
& Self-Adapting Linear & -- \\
& Algebra Algorithms~\cite{demmel:2005:selflinear} & -- \\
\bottomrule
\end{tabular}
\end{table}

\subsection{Pre-deployment Optimization}

In recent years, there have been significant efforts in automating the work to adapt an algorithm to  perform optimally on a particular platform. To some extent, the entire field of compiler optimization is dedicated to this goal, but recently there has been a strong push to move beyond the traditional application of pre-defined optimization steps according to a deterministic schedule and to embrace  learning-based techniques and search in order to explore the space of possible implementations to find a truly optimal one. 

A turning point for the field came with the advent of auto-tuning, first popularized by the ATLAS~\cite{WhaleyD98} and FFTW~\cite{FrigoJ98} projects in the late 90s. The high-level idea of auto-tuning is to explicitly explore a space of possible implementation choices to discover the one that works most efficiently on a particular architecture. The PetaBricks language pushed this idea all the way to the language level, allowing the programmer to explicitly provide implementation choices throughout the code, replacing all compiler heuristics with reinforcement learning to discover close to optimal implementations~\cite{Ansel:2009:petabricks}. 

Starting in the mid 2000s, there was also a realization that domain specific languages (DSLs) offered an important opportunity for automation. By eliminating a lot of the complexity of full-featured programming languages and exploiting domain specific representations, DSLs enabled aggressive symbolic manipulation of the computation, allowing the system to explore a much wider range of implementation strategies than what a human could possibly consider. Spiral~\cite{PuschelMSXJPVJ04} and the Tensor Contraction Engine (TCE)~\cite{BaumgartnerBCHHLNPRS02} were early examples of this approach. More recently, Halide has demonstrated the potential of this approach to bridge the ''ninja-gap'' by generating code that significantly outperforms expert-tuned codes with only a small amount of high-level guidance from the developer~\cite{Ragan-KelleyBAPDA13}.

Despite this successes, there is significant scope for advances in this direction. For example, how do we enable transfer learning, so that the $N$-th program can be optimized faster by leveraging learned optimizations from the previous ($N-1$) programs? Could a system learn new optimization strategies by analyzing a corpus of existing hand-optimized programs? Could learning help reduce the cost of developing high-performance DSLs, for example by reducing the need for custom heuristics or optimizations?

\subsection{Post-deployment Maintenance}

One of the most important maintenance tasks today is 
the repair of bugs and vulnerabilities.
Fixing software bugs is currently an entirely manual process. A programmer must diagnose, isolate, and correct the bug. While the bug remains in place, it can impair program behavior or even open up security vulnerabilities. Recent research has demonstrated the feasibility of automating many aspects of repair. Two early systems include ClearView~\cite{ClearView} and GenProg~\cite{WeimerNGF09}. ClearView uses learned invariants that characterize correct execution to generate patches (which can be applied to a running program) that repair a range of execution integrity defects. GenProg uses genetic programming to search for patches for defects exposed by input/output pairs. More recently, Fan Long and  Martin Rinard have pioneered machine learning mechanisms for automatically generating correct patches for large software systems~\cite{CodePhage,Genesis}. 

This recent research has highlighted the importance of learning and statistical techniques for program repair. At a fundamental level, program repair is an underdetermined problem, so a repair system must be able to select among all the possible patches that eliminate the symptoms of the bug to select the one that actually eliminates the bug without introducing other undesired behaviors. This can be done by automatically learning invariants that characterize correct behavior so that the generated patch can be required to maintain these invariants~\cite{ClearView}, by using machine learning over a large corpus of real bug fixes to build a model of successful patches~\cite{Prophet}, using change statistics from past human patches~\cite{HDP}, or even leveraging a large corpus of bug fixes to learn how to generate successful patches~\cite{Genesis}. 

Other successful program repair techniques focus on specific classes of defects such as incorrect conditionals~\cite{Nopol,ACS}. Here constraint solving can play an important role~\cite{SPR,SemFix,Angelix,Qlose,AngelicDebugging}. Automating repetitive source code edits can also eliminate or correct errors introduced by developer mistakes when working with similar code patterns~\cite{SYDIT,Lase,FixMeUp,REFAZER}. Code transfer, potentially augmented with machine learning to find appropriate code to transfer, can automatically work with code across multiple applications to eliminate defects and security vulnerabilities~\cite{MuScalpel,CodePhage,CodeCarbonCopy}.

These demonstrated techniques lay out the initial case for the feasibility of automated bug detection and correction, autonomously without programmer involvement. Many existing successful techniques focus largely on surviving execution integrity bugs (bugs that can cause crashes or open up security vulnerabilities). Future directions include the additional incorporation of machine learning to enhance current latent bug detection techniques and to generate more sophisticated corrections for larger classes of bugs.

In addition to bug fixing, there are a number of other post-deployment maintenance tasks that could benefit from learning. In general these fall into the category of bit-rot prevention, and include, for example, upgrading to new versions of APIs and web-services, porting to new platforms, such as new cloud or mobile environments, or specializing code for particular uses. 
We envision the eventual development of systems that continuously monitor program execution, incorporate user feedback, and learn from large code repositories to deliver a system of autonomous and continuous program correction and improvement~\cite{DIODE,CodePhage}.

\section{The Interplay Between Pillars}
\label{sect:interplay}

Systems for machine programming will most likely be composed of a set of tools each of which focuses on a particular pillar.  We anticipate that, in most cases, an individual tool cannot be fully understood in terms of a single pillar.  The machine programming problem is multifaceted and issues concerning one pillar will inevitably impact the other pillars.  Hence, we need to understand machine programming systems in terms of the interplay between the three pillars.

We expect this interplay to expose a tension between features of a tool that are supportive of the needs of any given pillar and those that are disruptive to the needs of the other pillars.  The challenge in designing a machine programming system is to understand this interplay and reach an effective resolution of those tensions.  As an example to explore this interplay, consider verified lifting~\cite{kamil:2016:VLS}.

Verified lifting tools input code written in one language, translate the code into a new language, and then formally verify that the new code produces results that are consistent with the original code.  The prototypical example~\cite{kamil:2016:VLS}  takes stencil codes written in an imperative language, translates them into a modern DSL such as Halide, and then uses theorem proving technology to verify that the original and generated DSL codes are functionally the same.  The newer code defines an abstract representation of the problem that can adapt onto a wide range of computer systems.  Therefore, we see that the verified lifting problem is primarily used to support the \emph{adaptation} pillar.

Verified lifting, however, goes well beyond the adaptation pillar.  Consider the early steps in the verified lifting problem.  A verified lifting tool must first understand the problem as represented in the input code.  It discovers the intent of the program and produces an internal high-level representation of the problem often in mathematical or functional terms.  This phase of the verified lifting process is firmly grounded in the \emph{intention} pillar.  From the high-level representation of the original problem, the verified lifting system can explore a range of algorithms appropriate to the target language; therefore working within the \emph{invention} pillar.  It then synthesizes the new code (the \emph{adaption} pillar) and verifies that it is consistent with the high level representation of the problem. Hence, a verified lifting tool, while nominally focused on adaption, touches, in a supportive way, all three pillars.  

It is important, however, to consider ways that a tool disrupts analysis within the different pillars. For example, when verified lifting translates low-level code into a compact representation in a DSL it is making the intent behind the code more apparent. Yet, the transformation can also interfere with other tools at the intention layer. For example, if the lifting transformation is not careful to preserve variable names, it may hamper the performance of intention layer tools that focus on names in the code to estimate whether a piece of code is relevant for a particular task. In general, when adaptation layer tools modify code, it is important to think about how the change may impact the ability of intention and invention layer tools to use that code.


While we have discussed the interplay between pillars in terms of just one machine based programming technique (verified lifting) we expect this complex interplay to be a common feature of machine programming systems.  As researchers in machine program apply the three pillars in their own research, it is essential to consider the interplay between the three pillars and how this interplay is supportive or disruptive to the overall programming process.

\section{Data}

Nearly all machine programming systems require \emph{data} to drive their algorithms. More specifically, every \emph{Research Area} listed in Tables~\ref{tab:intention}, \ref{tab:invention}, and \ref{tab:adaptation} requires data (in some form) to function properly. The data required by these subdomains comes in a variety of forms (e.g., code, input/output examples, DSLs, etc.), but is ever-present. This dependency on data makes it essential that we consider the open problems and emerging uses around data when reasoning about machine programming and the systems that implement it.



The various approaches to address the three machine programming pillars have different needs in terms of the type and size of data they require. Moreover, there is a wide spectrum in terms of the quality of the data that a project might use. We discuss some of these emerging data uses and issues for the remainder of this section.

\paragraph{Code Repositories}
Large version control repositories, such as GitHub, offer the promise of access to full revision histories for all the code necessary to build and run a project, as well as its accompanying documentation. The code available in these public repositories has grown  exponentially over the last several years and show no indication of stopping. Many projects in these repositories have long commit histories with detailed commit logs which could be of notable value to machine programming systems~\cite{vechev:2016:pbigcode}. However, recent analysis of public repositories has shown that a large fraction of the projects are duplicates, making a significant portion of the data less useful~\cite{LopesMMSYZSV17}. 

One use of code repositories is to use their version control histories to identify code changes that correspond to the introduction of performance or correctness bugs. This type of data utilization has been explored to train models for program repair~\cite{Prophet, mejbah:2017:autoperf}. Additionally, the presence of complete codebases makes it possible to run whole program analyses on the code. In some contexts, it has been shown that augmenting the code with features discovered from program analysis can help train more effective models~\cite{Raychev:2015:PPP}. However, complete codebases may not always be available, and running whole program analysis may not always be feasible.

\paragraph{Incomplete and Synthetic Code + Natural Language}
Sources such as stackoverflow provide a wealth of information beyond code, which can be used to correlate code and natural language. There has been some work in the community in extracting information from code that comes in the form of code snippets like those usually found in stackoverflow. This requires assembling information gathered from multiple different snippets into a coherent model of the behavior of a code component~\cite{PelegSYY13,Mishne:2012:TSC}. There have even been some efforts aimed at extracting code from video tutorials, which offers the possibility of correlating the code with the accompanying narration~\cite{Yadid:2016:ECP}. In some contexts, the data needed to train a model does not even have to come from real code; synthetic data generated from random combinations of components can be useful, as demonstrated by DeepCoder~\cite{balog:2017:iclr}. 

\paragraph{Data Privacy}
One of the open issues of machine programming data is that of privacy. In the context of code, machine programming systems will eventually have to work with and protect intellectual property as well as software licensing agreements. As we move toward a future where data will be more openly shared, used, and traded, new models and tools for secure and privacy-preserving exchange will become increasingly important~\cite{popa:2011:sosp, zheng:2017:opaque}. In the case of models learned from code, there are important open cyclic questions surrounding the copyright status of code generated from models trained from copyrighted code. 

\paragraph{Lifecycle Management}
Machine programming systems will require lifecycle management practices, similar in scope to those used in traditional software engineering. Much of this is due to the need to fulfill the goals of the adaptation pillar. These lifecycle management efforts will be long-lasting and will require support for continued monitoring and improvement around changing software needs and advances in an increasingly complex and heterogeneous hardware ecosystem. A significant portion of this lifecycle management will be centered on managing the data that are required for such an adaptive machine programming system, as these data will help ensure the stability and maturity of the system. As machine programming systems evolve so will the data they ingest to baseline and advance the system. Because of this, proper data management is likely to be a key enabler to calibrate any machine programming system's lifecycle.

\section{Conclusion}

In the post Dennard scaling world, where performance comes from architectural innovation rather than increased transistor count with constant power density, hardware complexity will only increase. Heterogeneous computing will become more widely used and more diverse than it is today.  Over the next several years, specialized accelerators will play an increasingly important role in the hardware platforms we depend on. At the same time, the nature of programming is changing. Instead of computer scientists trained in the low level details of how to map algorithms onto hardware, programmers are more likely to come from a broad range of academic and business backgrounds. Moreover, rather than programming in low level languages that interface almost directly to hardware, programmers are more likely to use higher level abstractions and scripting languages. This will fundamentally change how we write software. We believe this change is already well underway. 

We envision a future where computers will participate directly in the creation of software, which we call \emph{machine programming}. This paper presents a framework to organize work on this problem. We call this framework \emph{the three pillars of machine programming}. The three pillars are intention, invention, and adaptation. 

\emph{Intention} focuses on the interface between the human and the machine programming system; i.e., techniques to discover what a program needs to do from input that is natural to express by the human. A system grounded in the intention pillar meets human programmers on their terms rather than forcing them to express code in computer/hardware notations. \emph{Invention} emphasizes machine systems that create and refine algorithms or the core hardware and software building blocks from which systems are built. \emph{Adaptation} focuses on ML-based tools that help software adapt to changing conditions; whether they are bugs or vulnerabilities found in an application or new hardware systems.  

Data is at the foundation of the modern renaissance in artificial intelligence (AI). Without vast amounts of data, it is unlikely that AI would have had significant impact outside specialized academic circles. In this paper, we explored the impact of data as it pertains to machine programming. Software repositories in systems such as GitHub and the vast amount of software embedded in countless webpages is the raw material that will likely support a large majority of the emergence of machine programming.

Finally, there are numerous open problems that must be solved to make machine programming a practical reality. We outlined some of these open problems in this paper. It will take a large community of researchers years of hard work to solve this problem.  If we can agree on a conceptual framework to organize this research, it will help us advance the field and more quickly bring us to a world where everyone programs computers; on human-terms with machine systems handling the low level details of finding the right algorithm for the right hardware to solve the right problem.



%

\balance

\bibliographystyle{ACM-Reference-Format}
\bibliography{ml}


\begin{thebibliography}{84}


\ifx \showCODEN    \undefined \def \showCODEN     #1{\unskip}     \fi
\ifx \showDOI      \undefined \def \showDOI       #1{#1}\fi
\ifx \showISBNx    \undefined \def \showISBNx     #1{\unskip}     \fi
\ifx \showISBNxiii \undefined \def \showISBNxiii  #1{\unskip}     \fi
\ifx \showISSN     \undefined \def \showISSN      #1{\unskip}     \fi
\ifx \showLCCN     \undefined \def \showLCCN      #1{\unskip}     \fi
\ifx \shownote     \undefined \def \shownote      #1{#1}          \fi
\ifx \showarticletitle \undefined \def \showarticletitle #1{#1}   \fi
\ifx \showURL      \undefined \def \showURL       {\relax}        \fi
\providecommand\bibfield[2]{#2}
\providecommand\bibinfo[2]{#2}
\providecommand\natexlab[1]{#1}
\providecommand\showeprint[2][]{arXiv:#2}

\bibitem[\protect\citeauthoryear{Alam, Gottschlich, and Muzahid}{Alam
  et~al\mbox{.}}{2017}]%
        {mejbah:2017:autoperf}
\bibfield{author}{\bibinfo{person}{Mohammad Mejbah~Ul Alam},
  \bibinfo{person}{Justin Gottschlich}, {and} \bibinfo{person}{Abdullah
  Muzahid}.} \bibinfo{year}{2017}\natexlab{}.
\newblock \bibinfo{booktitle}{\emph{{AutoPerf: A Generalized Zero-Positive
  Learning System to Detect Software Performance Anomalies}}}.
\newblock \bibinfo{type}{{T}echnical {R}eport}.
\newblock
\urldef\tempurl%
\url{http://arxiv.org/abs/1709.07536/}
\showURL{%
\tempurl}


\bibitem[\protect\citeauthoryear{Alam and Muzahid}{Alam and Muzahid}{2016}]%
        {alam:2016:act}
\bibfield{author}{\bibinfo{person}{Mohammad Mejbah~ul Alam} {and}
  \bibinfo{person}{Abdullah Muzahid}.} \bibinfo{year}{2016}\natexlab{}.
\newblock \showarticletitle{Production-run Software Failure Diagnosis via
  Adaptive Communication Tracking}. In \bibinfo{booktitle}{\emph{Proceedings of
  the 43rd International Symposium on Computer Architecture}}
  \emph{(\bibinfo{series}{ISCA '16})}. \bibinfo{publisher}{IEEE Press},
  \bibinfo{address}{Piscataway, NJ, USA}, \bibinfo{pages}{354--366}.
\newblock
\showISBNx{978-1-4673-8947-1}
\urldef\tempurl%
\url{https://doi.org/10.1109/ISCA.2016.39}
\showDOI{\tempurl}


\bibitem[\protect\citeauthoryear{Albarghouthi, Gulwani, and
  Kincaid}{Albarghouthi et~al\mbox{.}}{2013}]%
        {AlbarghouthiGK13}
\bibfield{author}{\bibinfo{person}{Aws Albarghouthi}, \bibinfo{person}{Sumit
  Gulwani}, {and} \bibinfo{person}{Zachary Kincaid}.}
  \bibinfo{year}{2013}\natexlab{}.
\newblock \showarticletitle{Recursive Program Synthesis}. In
  \bibinfo{booktitle}{\emph{Computer Aided Verification - 25th International
  Conference, {CAV} 2013, Saint Petersburg, Russia, July 13-19, 2013.
  Proceedings}}. \bibinfo{pages}{934--950}.
\newblock
\urldef\tempurl%
\url{https://doi.org/10.1007/978-3-642-39799-8_67}
\showDOI{\tempurl}


\bibitem[\protect\citeauthoryear{Alur, Bod{\'{\i}}k, Dallal, Fisman, Garg,
  Juniwal, Kress{-}Gazit, Madhusudan, Martin, Raghothaman, Saha, Seshia, Singh,
  Solar{-}Lezama, Torlak, and Udupa}{Alur et~al\mbox{.}}{2015a}]%
        {sygus15}
\bibfield{author}{\bibinfo{person}{Rajeev Alur}, \bibinfo{person}{Rastislav
  Bod{\'{\i}}k}, \bibinfo{person}{Eric Dallal}, \bibinfo{person}{Dana Fisman},
  \bibinfo{person}{Pranav Garg}, \bibinfo{person}{Garvit Juniwal},
  \bibinfo{person}{Hadas Kress{-}Gazit}, \bibinfo{person}{P. Madhusudan},
  \bibinfo{person}{Milo M.~K. Martin}, \bibinfo{person}{Mukund Raghothaman},
  \bibinfo{person}{Shambwaditya Saha}, \bibinfo{person}{Sanjit~A. Seshia},
  \bibinfo{person}{Rishabh Singh}, \bibinfo{person}{Armando Solar{-}Lezama},
  \bibinfo{person}{Emina Torlak}, {and} \bibinfo{person}{Abhishek Udupa}.}
  \bibinfo{year}{2015}\natexlab{a}.
\newblock \showarticletitle{Syntax-Guided Synthesis}.
\newblock In \bibinfo{booktitle}{\emph{Dependable Software Systems
  Engineering}}. \bibinfo{pages}{1--25}.
\newblock
\urldef\tempurl%
\url{https://doi.org/10.3233/978-1-61499-495-4-1}
\showDOI{\tempurl}


\bibitem[\protect\citeauthoryear{Alur, Cern{\'{y}}, and Radhakrishna}{Alur
  et~al\mbox{.}}{2015b}]%
        {AlurCR15}
\bibfield{author}{\bibinfo{person}{Rajeev Alur}, \bibinfo{person}{Pavol
  Cern{\'{y}}}, {and} \bibinfo{person}{Arjun Radhakrishna}.}
  \bibinfo{year}{2015}\natexlab{b}.
\newblock \showarticletitle{Synthesis Through Unification}. In
  \bibinfo{booktitle}{\emph{Computer Aided Verification - 27th International
  Conference, {CAV} 2015, San Francisco, CA, USA, July 18-24, 2015,
  Proceedings, Part {II}}}. \bibinfo{pages}{163--179}.
\newblock
\urldef\tempurl%
\url{https://doi.org/10.1007/978-3-319-21668-3_10}
\showDOI{\tempurl}


\bibitem[\protect\citeauthoryear{Ansel, Chan, Wong, Olszewski, Zhao, Edelman,
  and Amarasinghe}{Ansel et~al\mbox{.}}{2009}]%
        {Ansel:2009:petabricks}
\bibfield{author}{\bibinfo{person}{Jason Ansel}, \bibinfo{person}{Cy Chan},
  \bibinfo{person}{Yee~Lok Wong}, \bibinfo{person}{Marek Olszewski},
  \bibinfo{person}{Qin Zhao}, \bibinfo{person}{Alan Edelman}, {and}
  \bibinfo{person}{Saman Amarasinghe}.} \bibinfo{year}{2009}\natexlab{}.
\newblock \showarticletitle{PetaBricks: A Language and Compiler for Algorithmic
  Choice}. In \bibinfo{booktitle}{\emph{Proceedings of the 30th ACM SIGPLAN
  Conference on Programming Language Design and Implementation}}
  \emph{(\bibinfo{series}{PLDI '09})}. \bibinfo{publisher}{ACM},
  \bibinfo{address}{New York, NY, USA}, \bibinfo{pages}{38--49}.
\newblock
\showISBNx{978-1-60558-392-1}
\urldef\tempurl%
\url{https://doi.org/10.1145/1542476.1542481}
\showDOI{\tempurl}


\bibitem[\protect\citeauthoryear{Ansel, Kamil, Veeramachaneni, Ragan-Kelley,
  Bosboom, O'Reilly, and Amarasinghe}{Ansel et~al\mbox{.}}{2014}]%
        {ansel:2014:opentuner}
\bibfield{author}{\bibinfo{person}{Jason Ansel}, \bibinfo{person}{Shoaib
  Kamil}, \bibinfo{person}{Kalyan Veeramachaneni}, \bibinfo{person}{Jonathan
  Ragan-Kelley}, \bibinfo{person}{Jeffrey Bosboom}, \bibinfo{person}{Una-May
  O'Reilly}, {and} \bibinfo{person}{Saman Amarasinghe}.}
  \bibinfo{year}{2014}\natexlab{}.
\newblock \showarticletitle{OpenTuner: An Extensible Framework for Program
  Autotuning}. In \bibinfo{booktitle}{\emph{Proceedings of the 23rd
  International Conference on Parallel Architectures and Compilation}}
  \emph{(\bibinfo{series}{PACT '14})}. \bibinfo{publisher}{ACM},
  \bibinfo{address}{New York, NY, USA}, \bibinfo{pages}{303--316}.
\newblock
\showISBNx{978-1-4503-2809-8}
\urldef\tempurl%
\url{https://doi.org/10.1145/2628071.2628092}
\showDOI{\tempurl}


\bibitem[\protect\citeauthoryear{Backus, Beeber, Best, Goldberg, Haibt,
  Herrick, Nelson, Sayre, Sheridan, Stern, Ziller, Hughes, and Nutt}{Backus
  et~al\mbox{.}}{1957}]%
        {Backus57}
\bibfield{author}{\bibinfo{person}{J.~W. Backus}, \bibinfo{person}{R.~J.
  Beeber}, \bibinfo{person}{S. Best}, \bibinfo{person}{R. Goldberg},
  \bibinfo{person}{L.~M. Haibt}, \bibinfo{person}{H.~L. Herrick},
  \bibinfo{person}{R.~A. Nelson}, \bibinfo{person}{D. Sayre},
  \bibinfo{person}{P.~B. Sheridan}, \bibinfo{person}{H. Stern},
  \bibinfo{person}{I. Ziller}, \bibinfo{person}{R.~A. Hughes}, {and}
  \bibinfo{person}{R. Nutt}.} \bibinfo{year}{1957}\natexlab{}.
\newblock \showarticletitle{The FORTRAN Automatic Coding System}. In
  \bibinfo{booktitle}{\emph{Papers Presented at the February 26-28, 1957,
  Western Joint Computer Conference: Techniques for Reliability}}
  \emph{(\bibinfo{series}{IRE-AIEE-ACM '57 (Western)})}.
  \bibinfo{publisher}{ACM}, \bibinfo{address}{New York, NY, USA},
  \bibinfo{pages}{188--198}.
\newblock
\urldef\tempurl%
\url{https://doi.org/10.1145/1455567.1455599}
\showDOI{\tempurl}


\bibitem[\protect\citeauthoryear{Balog, Gaunt, Brockschmidt, Nowozin, and
  Tarlow}{Balog et~al\mbox{.}}{2017}]%
        {balog:2017:iclr}
\bibfield{author}{\bibinfo{person}{Matej Balog}, \bibinfo{person}{Alexander~L.
  Gaunt}, \bibinfo{person}{Marc Brockschmidt}, \bibinfo{person}{Sebastian
  Nowozin}, {and} \bibinfo{person}{Daniel Tarlow}.}
  \bibinfo{year}{2017}\natexlab{}.
\newblock \showarticletitle{DeepCoder: Learning to Write Programs}.
\newblock \bibinfo{journal}{\emph{ICLR}} (\bibinfo{year}{2017}).
\newblock
\urldef\tempurl%
\url{https://arxiv.org/abs/1611.01989}
\showURL{%
\tempurl}


\bibitem[\protect\citeauthoryear{Barr, Harman, Jia, Marginean, and Petke}{Barr
  et~al\mbox{.}}{2015}]%
        {MuScalpel}
\bibfield{author}{\bibinfo{person}{Earl~T Barr}, \bibinfo{person}{Mark Harman},
  \bibinfo{person}{Yue Jia}, \bibinfo{person}{Alexandru Marginean}, {and}
  \bibinfo{person}{Justyna Petke}.} \bibinfo{year}{2015}\natexlab{}.
\newblock \showarticletitle{Automated software transplantation}. In
  \bibinfo{booktitle}{\emph{Proceedings of the 2015 International Symposium on
  Software Testing and Analysis}}. ACM, \bibinfo{pages}{257--269}.
\newblock


\bibitem[\protect\citeauthoryear{Barthe, Crespo, Gulwani, Kunz, and
  Marron}{Barthe et~al\mbox{.}}{2013}]%
        {BartheCKGM13}
\bibfield{author}{\bibinfo{person}{Gilles Barthe}, \bibinfo{person}{Juan~Manuel
  Crespo}, \bibinfo{person}{Sumit Gulwani}, \bibinfo{person}{C{\'{e}}sar Kunz},
  {and} \bibinfo{person}{Mark Marron}.} \bibinfo{year}{2013}\natexlab{}.
\newblock \showarticletitle{From relational verification to {SIMD} loop
  synthesis}. In \bibinfo{booktitle}{\emph{{ACM} {SIGPLAN} Symposium on
  Principles and Practice of Parallel Programming, PPoPP '13, Shenzhen, China,
  February 23-27, 2013}}. \bibinfo{pages}{123--134}.
\newblock
\urldef\tempurl%
\url{https://doi.org/10.1145/2442516.2442529}
\showDOI{\tempurl}


\bibitem[\protect\citeauthoryear{Baumgartner, Bernholdt, Cociorva, Harrison,
  Hirata, Lam, Nooijen, Pitzer, Ramanujam, and Sadayappan}{Baumgartner
  et~al\mbox{.}}{2002}]%
        {BaumgartnerBCHHLNPRS02}
\bibfield{author}{\bibinfo{person}{Gerald Baumgartner},
  \bibinfo{person}{David~E. Bernholdt}, \bibinfo{person}{Daniel Cociorva},
  \bibinfo{person}{Robert~J. Harrison}, \bibinfo{person}{So Hirata},
  \bibinfo{person}{Chi{-}Chung Lam}, \bibinfo{person}{Marcel Nooijen},
  \bibinfo{person}{Russell~M. Pitzer}, \bibinfo{person}{J. Ramanujam}, {and}
  \bibinfo{person}{P. Sadayappan}.} \bibinfo{year}{2002}\natexlab{}.
\newblock \showarticletitle{A high-level approach to synthesis of
  high-performance codes for quantum chemistry}. In
  \bibinfo{booktitle}{\emph{Proceedings of the 2002 {ACM/IEEE} conference on
  Supercomputing, Baltimore, Maryland, USA, November 16-22, 2002, {CD-ROM}}}.
  \bibinfo{pages}{33:1--33:10}.
\newblock
\urldef\tempurl%
\url{https://doi.org/10.1109/SC.2002.10056}
\showDOI{\tempurl}


\bibitem[\protect\citeauthoryear{Becker and Gottschlich}{Becker and
  Gottschlich}{2017}]%
        {becker:2017:ai}
\bibfield{author}{\bibinfo{person}{Kory Becker} {and} \bibinfo{person}{Justin
  Gottschlich}.} \bibinfo{year}{2017}\natexlab{}.
\newblock \showarticletitle{{AI} Programmer: Autonomously Creating Software
  Programs Using Genetic Algorithms}.
\newblock \bibinfo{journal}{\emph{CoRR}}  \bibinfo{volume}{abs/1709.05703}
  (\bibinfo{year}{2017}).
\newblock
\showeprint[arxiv]{1709.05703}
\urldef\tempurl%
\url{http://arxiv.org/abs/1709.05703}
\showURL{%
\tempurl}


\bibitem[\protect\citeauthoryear{Bielik, Raychev, and Vechev}{Bielik
  et~al\mbox{.}}{2016}]%
        {BielikRV16}
\bibfield{author}{\bibinfo{person}{Pavol Bielik}, \bibinfo{person}{Veselin
  Raychev}, {and} \bibinfo{person}{Martin~T. Vechev}.}
  \bibinfo{year}{2016}\natexlab{}.
\newblock \showarticletitle{{PHOG:} Probabilistic Model for Code}. In
  \bibinfo{booktitle}{\emph{Proceedings of the 33nd International Conference on
  Machine Learning, {ICML} 2016, New York City, NY, USA, June 19-24, 2016}}.
  \bibinfo{pages}{2933--2942}.
\newblock
\urldef\tempurl%
\url{http://jmlr.org/proceedings/papers/v48/bielik16.html}
\showURL{%
\tempurl}


\bibitem[\protect\citeauthoryear{Cai, Shin, and Song}{Cai
  et~al\mbox{.}}{2017}]%
        {CaiSS17}
\bibfield{author}{\bibinfo{person}{Jonathon Cai}, \bibinfo{person}{Richard
  Shin}, {and} \bibinfo{person}{Dawn Song}.} \bibinfo{year}{2017}\natexlab{}.
\newblock \showarticletitle{Making Neural Programming Architectures Generalize
  via Recursion}.
\newblock \bibinfo{journal}{\emph{CoRR}}  \bibinfo{volume}{abs/1704.06611}
  (\bibinfo{year}{2017}).
\newblock
\showeprint[arxiv]{1704.06611}
\urldef\tempurl%
\url{http://arxiv.org/abs/1704.06611}
\showURL{%
\tempurl}


\bibitem[\protect\citeauthoryear{Chandra, Torlak, Barman, and Bodik}{Chandra
  et~al\mbox{.}}{2011}]%
        {AngelicDebugging}
\bibfield{author}{\bibinfo{person}{Satish Chandra}, \bibinfo{person}{Emina
  Torlak}, \bibinfo{person}{Shaon Barman}, {and} \bibinfo{person}{Rastislav
  Bodik}.} \bibinfo{year}{2011}\natexlab{}.
\newblock \showarticletitle{Angelic Debugging}. In
  \bibinfo{booktitle}{\emph{Proceedings of the 33rd International Conference on
  Software Engineering}} \emph{(\bibinfo{series}{ICSE '11'})}.
\newblock


\bibitem[\protect\citeauthoryear{Chaudhuri and Solar{-}Lezama}{Chaudhuri and
  Solar{-}Lezama}{2010}]%
        {ChaudhuriS10}
\bibfield{author}{\bibinfo{person}{Swarat Chaudhuri} {and}
  \bibinfo{person}{Armando Solar{-}Lezama}.} \bibinfo{year}{2010}\natexlab{}.
\newblock \showarticletitle{Smooth interpretation}. In
  \bibinfo{booktitle}{\emph{Proceedings of the 2010 {ACM} {SIGPLAN} Conference
  on Programming Language Design and Implementation, {PLDI} 2010, Toronto,
  Ontario, Canada, June 5-10, 2010}}. \bibinfo{pages}{279--291}.
\newblock
\urldef\tempurl%
\url{https://doi.org/10.1145/1806596.1806629}
\showDOI{\tempurl}


\bibitem[\protect\citeauthoryear{Chen, Liu, and Song}{Chen
  et~al\mbox{.}}{2017a}]%
        {chen:2017:example}
\bibfield{author}{\bibinfo{person}{Xinyun Chen}, \bibinfo{person}{Chang Liu},
  {and} \bibinfo{person}{Dawn Song}.} \bibinfo{year}{2017}\natexlab{a}.
\newblock \showarticletitle{Towards Synthesizing Complex Programs from
  Input-Output Examples}.
\newblock \bibinfo{journal}{\emph{CoRR}}  \bibinfo{volume}{abs/1706.01284}
  (\bibinfo{year}{2017}).
\newblock
\showeprint[arxiv]{1706.01284}
\urldef\tempurl%
\url{http://arxiv.org/abs/1706.01284}
\showURL{%
\tempurl}


\bibitem[\protect\citeauthoryear{Chen, Liu, and Song}{Chen
  et~al\mbox{.}}{2017b}]%
        {chen:2017:tree}
\bibfield{author}{\bibinfo{person}{Xinyun Chen}, \bibinfo{person}{Chang Liu},
  {and} \bibinfo{person}{Dawn Song}.} \bibinfo{year}{2017}\natexlab{b}.
\newblock \showarticletitle{Tree-to-tree Neural Networks for Program
  Translation}.
\newblock \bibinfo{journal}{\emph{CoRR}}  \bibinfo{volume}{abs/1712.01208}
  (\bibinfo{year}{2017}).
\newblock
\urldef\tempurl%
\url{https://arxiv.org/abs/1712.01208}
\showURL{%
\tempurl}


\bibitem[\protect\citeauthoryear{Cypher, Halbert, Kurlander, Lieberman,
  Maulsby, Myers, and Turransky}{Cypher et~al\mbox{.}}{1993}]%
        {cypher:1993:wwid}
\bibfield{editor}{\bibinfo{person}{Allen Cypher}, \bibinfo{person}{Daniel~C.
  Halbert}, \bibinfo{person}{David Kurlander}, \bibinfo{person}{Henry
  Lieberman}, \bibinfo{person}{David Maulsby}, \bibinfo{person}{Brad~A. Myers},
  {and} \bibinfo{person}{Alan Turransky}} (Eds.).
  \bibinfo{year}{1993}\natexlab{}.
\newblock \bibinfo{booktitle}{\emph{Watch What I Do: Programming by
  Demonstration}}.
\newblock \bibinfo{publisher}{MIT Press}, \bibinfo{address}{Cambridge, MA,
  USA}.
\newblock
\showISBNx{0-262-03213-9}


\bibitem[\protect\citeauthoryear{D'Antoni, Samanta, and Singh}{D'Antoni
  et~al\mbox{.}}{2016}]%
        {Qlose}
\bibfield{author}{\bibinfo{person}{Loris D'Antoni}, \bibinfo{person}{Roopsha
  Samanta}, {and} \bibinfo{person}{Rishabh Singh}.}
  \bibinfo{year}{2016}\natexlab{}.
\newblock \showarticletitle{Qlose: Program Repair with Quantitative
  Objectives}. In \bibinfo{booktitle}{\emph{Computer-Aided Verification
  (CAV)}}.
\newblock


\bibitem[\protect\citeauthoryear{Delaware, Pit{-}Claudel, Gross, and
  Chlipala}{Delaware et~al\mbox{.}}{2015}]%
        {DelawarePGC15}
\bibfield{author}{\bibinfo{person}{Benjamin Delaware},
  \bibinfo{person}{Cl{\'{e}}ment Pit{-}Claudel}, \bibinfo{person}{Jason Gross},
  {and} \bibinfo{person}{Adam Chlipala}.} \bibinfo{year}{2015}\natexlab{}.
\newblock \showarticletitle{Fiat: Deductive Synthesis of Abstract Data Types in
  a Proof Assistant}. In \bibinfo{booktitle}{\emph{Proceedings of the 42nd
  Annual {ACM} {SIGPLAN-SIGACT} Symposium on Principles of Programming
  Languages, {POPL} 2015, Mumbai, India, January 15-17, 2015}}.
  \bibinfo{pages}{689--700}.
\newblock
\urldef\tempurl%
\url{https://doi.org/10.1145/2676726.2677006}
\showDOI{\tempurl}


\bibitem[\protect\citeauthoryear{Demmel, Dongarra, Eijkhout, Fuentes, Petitet,
  Vuduc, Whaley, and Yelick}{Demmel et~al\mbox{.}}{2005}]%
        {demmel:2005:selflinear}
\bibfield{author}{\bibinfo{person}{J. Demmel}, \bibinfo{person}{J. Dongarra},
  \bibinfo{person}{V. Eijkhout}, \bibinfo{person}{E. Fuentes},
  \bibinfo{person}{A. Petitet}, \bibinfo{person}{R. Vuduc},
  \bibinfo{person}{R.~C. Whaley}, {and} \bibinfo{person}{K. Yelick}.}
  \bibinfo{year}{2005}\natexlab{}.
\newblock \showarticletitle{Self-Adapting Linear Algebra Algorithms and
  Software}.
\newblock \bibinfo{journal}{\emph{Proc. IEEE}} \bibinfo{volume}{93},
  \bibinfo{number}{2} (\bibinfo{date}{Feb} \bibinfo{year}{2005}),
  \bibinfo{pages}{293--312}.
\newblock
\showISSN{0018-9219}
\urldef\tempurl%
\url{https://doi.org/10.1109/JPROC.2004.840848}
\showDOI{\tempurl}


\bibitem[\protect\citeauthoryear{Durieux, Martinez, Monperrus, Sommerard, and
  Xuan}{Durieux et~al\mbox{.}}{2015}]%
        {Nopol}
\bibfield{author}{\bibinfo{person}{Thomas Durieux}, \bibinfo{person}{Matias
  Martinez}, \bibinfo{person}{Martin Monperrus}, \bibinfo{person}{Romain
  Sommerard}, {and} \bibinfo{person}{Jifeng Xuan}.}
  \bibinfo{year}{2015}\natexlab{}.
\newblock \showarticletitle{Automatic Repair of Real Bugs: An Experience Report
  on the Defects4J Dataset}.
\newblock \bibinfo{journal}{\emph{CoRR}}  \bibinfo{volume}{abs/1505.07002}
  (\bibinfo{year}{2015}).
\newblock
\urldef\tempurl%
\url{http://arxiv.org/abs/1505.07002}
\showURL{%
\tempurl}


\bibitem[\protect\citeauthoryear{Ellis and Gulwani}{Ellis and Gulwani}{2017}]%
        {EllisG17}
\bibfield{author}{\bibinfo{person}{Kevin Ellis} {and} \bibinfo{person}{Sumit
  Gulwani}.} \bibinfo{year}{2017}\natexlab{}.
\newblock \showarticletitle{Learning to Learn Programs from Examples: Going
  Beyond Program Structure}. In \bibinfo{booktitle}{\emph{Proceedings of the
  Twenty-Sixth International Joint Conference on Artificial Intelligence,
  {IJCAI} 2017, Melbourne, Australia, August 19-25, 2017}}.
  \bibinfo{pages}{1638--1645}.
\newblock
\urldef\tempurl%
\url{https://doi.org/10.24963/ijcai.2017/227}
\showDOI{\tempurl}


\bibitem[\protect\citeauthoryear{Ellis, Ritchie, Solar{-}Lezama, and
  Tenenbaum}{Ellis et~al\mbox{.}}{2017}]%
        {EllisRST17}
\bibfield{author}{\bibinfo{person}{Kevin Ellis}, \bibinfo{person}{Daniel
  Ritchie}, \bibinfo{person}{Armando Solar{-}Lezama}, {and}
  \bibinfo{person}{Joshua~B. Tenenbaum}.} \bibinfo{year}{2017}\natexlab{}.
\newblock \showarticletitle{Learning to Infer Graphics Programs from Hand-Drawn
  Images}.
\newblock \bibinfo{journal}{\emph{CoRR}}  \bibinfo{volume}{abs/1707.09627}
  (\bibinfo{year}{2017}).
\newblock
\showeprint[arxiv]{1707.09627}
\urldef\tempurl%
\url{http://arxiv.org/abs/1707.09627}
\showURL{%
\tempurl}


\bibitem[\protect\citeauthoryear{Esmaeilzadeh, Blem, Amant, Sankaralingam, and
  Burger}{Esmaeilzadeh et~al\mbox{.}}{2011}]%
        {esmae:2011:dennard}
\bibfield{author}{\bibinfo{person}{H. Esmaeilzadeh}, \bibinfo{person}{E. Blem},
  \bibinfo{person}{R.~S. Amant}, \bibinfo{person}{K. Sankaralingam}, {and}
  \bibinfo{person}{D. Burger}.} \bibinfo{year}{2011}\natexlab{}.
\newblock \showarticletitle{Dark silicon and the end of multicore scaling}. In
  \bibinfo{booktitle}{\emph{2011 38th Annual International Symposium on
  Computer Architecture (ISCA)}}. \bibinfo{pages}{365--376}.
\newblock
\showISSN{1063-6897}


\bibitem[\protect\citeauthoryear{Feser, Chaudhuri, and Dillig}{Feser
  et~al\mbox{.}}{2015}]%
        {FeserCD15}
\bibfield{author}{\bibinfo{person}{John~K. Feser}, \bibinfo{person}{Swarat
  Chaudhuri}, {and} \bibinfo{person}{Isil Dillig}.}
  \bibinfo{year}{2015}\natexlab{}.
\newblock \showarticletitle{Synthesizing data structure transformations from
  input-output examples}. In \bibinfo{booktitle}{\emph{Proceedings of the 36th
  {ACM} {SIGPLAN} Conference on Programming Language Design and Implementation,
  Portland, OR, USA, June 15-17, 2015}}. \bibinfo{pages}{229--239}.
\newblock
\urldef\tempurl%
\url{https://doi.org/10.1145/2737924.2737977}
\showDOI{\tempurl}


\bibitem[\protect\citeauthoryear{Frigo and Johnson}{Frigo and Johnson}{1998}]%
        {FrigoJ98}
\bibfield{author}{\bibinfo{person}{Matteo Frigo} {and}
  \bibinfo{person}{Steven~G. Johnson}.} \bibinfo{year}{1998}\natexlab{}.
\newblock \showarticletitle{{FFTW:} an adaptive software architecture for the
  {FFT}}. In \bibinfo{booktitle}{\emph{Proceedings of the 1998 {IEEE}
  International Conference on Acoustics, Speech and Signal Processing, {ICASSP}
  '98, Seattle, Washington, USA, May 12-15, 1998}}.
  \bibinfo{pages}{1381--1384}.
\newblock
\urldef\tempurl%
\url{https://doi.org/10.1109/ICASSP.1998.681704}
\showDOI{\tempurl}


\bibitem[\protect\citeauthoryear{Gaunt, Brockschmidt, Singh, Kushman, Kohli,
  Taylor, and Tarlow}{Gaunt et~al\mbox{.}}{2016}]%
        {GauntBSKKTT16}
\bibfield{author}{\bibinfo{person}{Alexander~L. Gaunt}, \bibinfo{person}{Marc
  Brockschmidt}, \bibinfo{person}{Rishabh Singh}, \bibinfo{person}{Nate
  Kushman}, \bibinfo{person}{Pushmeet Kohli}, \bibinfo{person}{Jonathan
  Taylor}, {and} \bibinfo{person}{Daniel Tarlow}.}
  \bibinfo{year}{2016}\natexlab{}.
\newblock \showarticletitle{TerpreT: {A} Probabilistic Programming Language for
  Program Induction}.
\newblock \bibinfo{journal}{\emph{CoRR}}  \bibinfo{volume}{abs/1608.04428}
  (\bibinfo{year}{2016}).
\newblock
\showeprint[arxiv]{1608.04428}
\urldef\tempurl%
\url{http://arxiv.org/abs/1608.04428}
\showURL{%
\tempurl}


\bibitem[\protect\citeauthoryear{Graves, Wayne, and Danihelka}{Graves
  et~al\mbox{.}}{2014}]%
        {GravesWD14}
\bibfield{author}{\bibinfo{person}{Alex Graves}, \bibinfo{person}{Greg Wayne},
  {and} \bibinfo{person}{Ivo Danihelka}.} \bibinfo{year}{2014}\natexlab{}.
\newblock \showarticletitle{Neural Turing Machines}.
\newblock \bibinfo{journal}{\emph{CoRR}}  \bibinfo{volume}{abs/1410.5401}
  (\bibinfo{year}{2014}).
\newblock
\showeprint[arxiv]{1410.5401}
\urldef\tempurl%
\url{http://arxiv.org/abs/1410.5401}
\showURL{%
\tempurl}


\bibitem[\protect\citeauthoryear{Gulwani}{Gulwani}{2011}]%
        {gulwani:2011:flashfill}
\bibfield{author}{\bibinfo{person}{Sumit Gulwani}.}
  \bibinfo{year}{2011}\natexlab{}.
\newblock \showarticletitle{Automating String Processing in Spreadsheets Using
  Input-output Examples}. In \bibinfo{booktitle}{\emph{Proceedings of the 38th
  Annual ACM SIGPLAN-SIGACT Symposium on Principles of Programming Languages}}
  \emph{(\bibinfo{series}{POPL '11})}. \bibinfo{publisher}{ACM},
  \bibinfo{address}{New York, NY, USA}, \bibinfo{pages}{317--330}.
\newblock
\showISBNx{978-1-4503-0490-0}
\urldef\tempurl%
\url{https://doi.org/10.1145/1926385.1926423}
\showDOI{\tempurl}


\bibitem[\protect\citeauthoryear{Gulwani, Jha, Tiwari, and Venkatesan}{Gulwani
  et~al\mbox{.}}{2011}]%
        {GulwaniJTV11}
\bibfield{author}{\bibinfo{person}{Sumit Gulwani}, \bibinfo{person}{Susmit
  Jha}, \bibinfo{person}{Ashish Tiwari}, {and} \bibinfo{person}{Ramarathnam
  Venkatesan}.} \bibinfo{year}{2011}\natexlab{}.
\newblock \showarticletitle{Synthesis of loop-free programs}. In
  \bibinfo{booktitle}{\emph{Proceedings of the 32nd {ACM} {SIGPLAN} Conference
  on Programming Language Design and Implementation, {PLDI} 2011, San Jose, CA,
  USA, June 4-8, 2011}}. \bibinfo{pages}{62--73}.
\newblock
\urldef\tempurl%
\url{https://doi.org/10.1145/1993498.1993506}
\showDOI{\tempurl}


\bibitem[\protect\citeauthoryear{Inala, Singh, and Solar{-}Lezama}{Inala
  et~al\mbox{.}}{2016}]%
        {Inala0S16}
\bibfield{author}{\bibinfo{person}{Jeevana~Priya Inala}, \bibinfo{person}{Rohit
  Singh}, {and} \bibinfo{person}{Armando Solar{-}Lezama}.}
  \bibinfo{year}{2016}\natexlab{}.
\newblock \showarticletitle{Synthesis of Domain Specific {CNF} Encoders for
  Bit-Vector Solvers}. In \bibinfo{booktitle}{\emph{Theory and Applications of
  Satisfiability Testing - {SAT} 2016 - 19th International Conference,
  Bordeaux, France, July 5-8, 2016, Proceedings}}. \bibinfo{pages}{302--320}.
\newblock
\urldef\tempurl%
\url{https://doi.org/10.1007/978-3-319-40970-2_19}
\showDOI{\tempurl}


\bibitem[\protect\citeauthoryear{Iyer, Konstas, Cheung, Krishnamurthy, and
  Zettlemoyer}{Iyer et~al\mbox{.}}{2017}]%
        {IyerKCKZ17}
\bibfield{author}{\bibinfo{person}{Srinivasan Iyer}, \bibinfo{person}{Ioannis
  Konstas}, \bibinfo{person}{Alvin Cheung}, \bibinfo{person}{Jayant
  Krishnamurthy}, {and} \bibinfo{person}{Luke Zettlemoyer}.}
  \bibinfo{year}{2017}\natexlab{}.
\newblock \showarticletitle{Learning a Neural Semantic Parser from User
  Feedback}. In \bibinfo{booktitle}{\emph{Proceedings of the 55th Annual
  Meeting of the Association for Computational Linguistics, {ACL} 2017,
  Vancouver, Canada, July 30 - August 4, Volume 1: Long Papers}}.
  \bibinfo{pages}{963--973}.
\newblock
\urldef\tempurl%
\url{https://doi.org/10.18653/v1/P17-1089}
\showDOI{\tempurl}


\bibitem[\protect\citeauthoryear{Kamil, Cheung, Itzhaky, and
  Solar-Lezama}{Kamil et~al\mbox{.}}{2016}]%
        {kamil:2016:VLS}
\bibfield{author}{\bibinfo{person}{Shoaib Kamil}, \bibinfo{person}{Alvin
  Cheung}, \bibinfo{person}{Shachar Itzhaky}, {and} \bibinfo{person}{Armando
  Solar-Lezama}.} \bibinfo{year}{2016}\natexlab{}.
\newblock \showarticletitle{Verified Lifting of Stencil Computations}. In
  \bibinfo{booktitle}{\emph{Proceedings of the 37th ACM SIGPLAN Conference on
  Programming Language Design and Implementation}} \emph{(\bibinfo{series}{PLDI
  '16})}. \bibinfo{publisher}{ACM}, \bibinfo{address}{New York, NY, USA},
  \bibinfo{pages}{711--726}.
\newblock
\showISBNx{978-1-4503-4261-2}
\urldef\tempurl%
\url{https://doi.org/10.1145/2908080.2908117}
\showDOI{\tempurl}


\bibitem[\protect\citeauthoryear{Kraska, Beutel, Chi, Dean, and
  Polyzotis}{Kraska et~al\mbox{.}}{2017}]%
        {kraska:2017:learned}
\bibfield{author}{\bibinfo{person}{Tim Kraska}, \bibinfo{person}{Alex Beutel},
  \bibinfo{person}{Ed~H. Chi}, \bibinfo{person}{Jeffrey Dean}, {and}
  \bibinfo{person}{Neoklis Polyzotis}.} \bibinfo{year}{2017}\natexlab{}.
\newblock \showarticletitle{The Case for Learned Index Structures}.
\newblock \bibinfo{journal}{\emph{CoRR}}  \bibinfo{volume}{abs/1712.01208}
  (\bibinfo{year}{2017}).
\newblock
\urldef\tempurl%
\url{https://arxiv.org/abs/1712.01208}
\showURL{%
\tempurl}


\bibitem[\protect\citeauthoryear{Le, Lo, and {Le Goues}}{Le
  et~al\mbox{.}}{2016}]%
        {HDP}
\bibfield{author}{\bibinfo{person}{Xuan{-}Bach~D. Le}, \bibinfo{person}{David
  Lo}, {and} \bibinfo{person}{Claire {Le Goues}}.}
  \bibinfo{year}{2016}\natexlab{}.
\newblock \showarticletitle{History Driven Program Repair}. In
  \bibinfo{booktitle}{\emph{{IEEE} 23rd International Conference on Software
  Analysis, Evolution, and Reengineering, {SANER} 2016, Suita, Osaka, Japan,
  March 14-18, 2016}}. \bibinfo{pages}{213--224}.
\newblock


\bibitem[\protect\citeauthoryear{Lei, Long, Barzilay, and Rinard}{Lei
  et~al\mbox{.}}{2013}]%
        {NL2P}
\bibfield{author}{\bibinfo{person}{Tao Lei}, \bibinfo{person}{Fan Long},
  \bibinfo{person}{Regina Barzilay}, {and} \bibinfo{person}{Martin~C. Rinard}.}
  \bibinfo{year}{2013}\natexlab{}.
\newblock \showarticletitle{From Natural Language Specifications to Program
  Input Parsers}. In \bibinfo{booktitle}{\emph{Proceedings of the 51st Annual
  Meeting of the Association for Computational Linguistics, {ACL} 2013, 4-9
  August 2013, Sofia, Bulgaria, Volume 1: Long Papers}}.
  \bibinfo{pages}{1294--1303}.
\newblock


\bibitem[\protect\citeauthoryear{Li and Malik}{Li and Malik}{2017}]%
        {li:2017:iclr}
\bibfield{author}{\bibinfo{person}{Ke~Li Li} {and} \bibinfo{person}{Jitendra
  Malik}.} \bibinfo{year}{2017}\natexlab{}.
\newblock \showarticletitle{Learning to Optimize}.
\newblock \bibinfo{journal}{\emph{ICLR}} (\bibinfo{year}{2017}).
\newblock
\urldef\tempurl%
\url{https://arxiv.org/abs/1606.01885}
\showURL{%
\tempurl}


\bibitem[\protect\citeauthoryear{Long, Amidon, and Rinard}{Long
  et~al\mbox{.}}{2017}]%
        {Genesis}
\bibfield{author}{\bibinfo{person}{Fan Long}, \bibinfo{person}{Peter Amidon},
  {and} \bibinfo{person}{Martin Rinard}.} \bibinfo{year}{2017}\natexlab{}.
\newblock \showarticletitle{Automatic inference of code transforms for patch
  generation}. In \bibinfo{booktitle}{\emph{Proceedings of the 2017 11th Joint
  Meeting on Foundations of Software Engineering, {ESEC/FSE} 2017, Paderborn,
  Germany, September 4-8, 2017}}. \bibinfo{pages}{727--739}.
\newblock


\bibitem[\protect\citeauthoryear{Long and Rinard}{Long and Rinard}{2015}]%
        {SPR}
\bibfield{author}{\bibinfo{person}{Fan Long} {and} \bibinfo{person}{Martin
  Rinard}.} \bibinfo{year}{2015}\natexlab{}.
\newblock \showarticletitle{Staged Program Repair with Condition Synthesis}. In
  \bibinfo{booktitle}{\emph{Proceedings of the 2015 10th Joint Meeting on
  Foundations of Software Engineering}} \emph{(\bibinfo{series}{ESEC/FSE
  2015})}.
\newblock


\bibitem[\protect\citeauthoryear{Long and Rinard}{Long and Rinard}{2016}]%
        {Prophet}
\bibfield{author}{\bibinfo{person}{Fan Long} {and} \bibinfo{person}{Martin
  Rinard}.} \bibinfo{year}{2016}\natexlab{}.
\newblock \showarticletitle{Automatic patch generation by learning correct
  code}. In \bibinfo{booktitle}{\emph{Proceedings of the 43rd Annual {ACM}
  {SIGPLAN-SIGACT} Symposium on Principles of Programming Languages, {POPL}
  2016, St. Petersburg, FL, USA, January 20 - 22, 2016}}.
  \bibinfo{pages}{298--312}.
\newblock


\bibitem[\protect\citeauthoryear{Lopes, Maj, Martins, Saini, Yang, Zitny,
  Sajnani, and Vitek}{Lopes et~al\mbox{.}}{2017}]%
        {LopesMMSYZSV17}
\bibfield{author}{\bibinfo{person}{Cristina~V. Lopes}, \bibinfo{person}{Petr
  Maj}, \bibinfo{person}{Pedro Martins}, \bibinfo{person}{Vaibhav Saini},
  \bibinfo{person}{Di Yang}, \bibinfo{person}{Jakub Zitny},
  \bibinfo{person}{Hitesh Sajnani}, {and} \bibinfo{person}{Jan Vitek}.}
  \bibinfo{year}{2017}\natexlab{}.
\newblock \showarticletitle{D{\'{e}}j{\`{a}}Vu: a map of code duplicates on
  GitHub}.
\newblock \bibinfo{journal}{\emph{{PACMPL}}} \bibinfo{volume}{1},
  \bibinfo{number}{{OOPSLA}} (\bibinfo{year}{2017}),
  \bibinfo{pages}{84:1--84:28}.
\newblock
\urldef\tempurl%
\url{https://doi.org/10.1145/3133908}
\showDOI{\tempurl}


\bibitem[\protect\citeauthoryear{Mechtaev, Yi, and Roychoudhury}{Mechtaev
  et~al\mbox{.}}{2016}]%
        {Angelix}
\bibfield{author}{\bibinfo{person}{Sergey Mechtaev}, \bibinfo{person}{Jooyong
  Yi}, {and} \bibinfo{person}{Abhik Roychoudhury}.}
  \bibinfo{year}{2016}\natexlab{}.
\newblock \showarticletitle{Angelix: scalable multiline program patch synthesis
  via symbolic analysis}. In \bibinfo{booktitle}{\emph{Proceedings of the 38th
  International Conference on Software Engineering, {ICSE} 2016, Austin, TX,
  USA, May 14-22, 2016}}. \bibinfo{pages}{691--701}.
\newblock


\bibitem[\protect\citeauthoryear{Meng, Kim, and McKinley}{Meng
  et~al\mbox{.}}{2011}]%
        {SYDIT}
\bibfield{author}{\bibinfo{person}{Na Meng}, \bibinfo{person}{Miryung Kim},
  {and} \bibinfo{person}{Kathryn~S. McKinley}.}
  \bibinfo{year}{2011}\natexlab{}.
\newblock \showarticletitle{Systematic Editing: Generating Program
  Transformations from an Example}. In \bibinfo{booktitle}{\emph{Proceedings of
  the 32Nd ACM SIGPLAN Conference on Programming Language Design and
  Implementation}} \emph{(\bibinfo{series}{PLDI '11'})}.
  \bibinfo{pages}{329--342}.
\newblock
\showISBNx{978-1-4503-0663-8}


\bibitem[\protect\citeauthoryear{Meng, Kim, and McKinley}{Meng
  et~al\mbox{.}}{2013}]%
        {Lase}
\bibfield{author}{\bibinfo{person}{Na Meng}, \bibinfo{person}{Miryung Kim},
  {and} \bibinfo{person}{Kathryn~S. McKinley}.}
  \bibinfo{year}{2013}\natexlab{}.
\newblock \showarticletitle{{LASE:} locating and applying systematic edits by
  learning from examples}. In \bibinfo{booktitle}{\emph{35th International
  Conference on Software Engineering, {ICSE} '13, San Francisco, CA, USA, May
  18-26, 2013}}. \bibinfo{pages}{502--511}.
\newblock


\bibitem[\protect\citeauthoryear{Mishne, Shoham, and Yahav}{Mishne
  et~al\mbox{.}}{2012}]%
        {Mishne:2012:TSC}
\bibfield{author}{\bibinfo{person}{Alon Mishne}, \bibinfo{person}{Sharon
  Shoham}, {and} \bibinfo{person}{Eran Yahav}.}
  \bibinfo{year}{2012}\natexlab{}.
\newblock \showarticletitle{Typestate-based Semantic Code Search over Partial
  Programs}. In \bibinfo{booktitle}{\emph{Proceedings of the ACM International
  Conference on Object Oriented Programming Systems Languages and
  Applications}} \emph{(\bibinfo{series}{OOPSLA '12})}.
  \bibinfo{publisher}{ACM}, \bibinfo{address}{New York, NY, USA},
  \bibinfo{pages}{997--1016}.
\newblock
\showISBNx{978-1-4503-1561-6}
\urldef\tempurl%
\url{https://doi.org/10.1145/2384616.2384689}
\showDOI{\tempurl}


\bibitem[\protect\citeauthoryear{Muggleton}{Muggleton}{1991}]%
        {muggleton:1991:ip}
\bibfield{author}{\bibinfo{person}{Stephen Muggleton}.}
  \bibinfo{year}{1991}\natexlab{}.
\newblock \bibinfo{booktitle}{\emph{Inductive logic programming}}.
  Vol.~\bibinfo{volume}{8}.
\newblock 295--318 pages.
\newblock
\showISSN{1882-7055}
\urldef\tempurl%
\url{https://doi.org/10.1007/BF03037089}
\showDOI{\tempurl}


\bibitem[\protect\citeauthoryear{Murali, Chaudhuri, and Jermaine}{Murali
  et~al\mbox{.}}{2017}]%
        {vijay:2017:bayou}
\bibfield{author}{\bibinfo{person}{Vijayaraghavan Murali},
  \bibinfo{person}{Swarat Chaudhuri}, {and} \bibinfo{person}{Chris Jermaine}.}
  \bibinfo{year}{2017}\natexlab{}.
\newblock \showarticletitle{Bayesian Sketch Learning for Program Synthesis}.
\newblock \bibinfo{journal}{\emph{CoRR}}  \bibinfo{volume}{abs/1703.05698}
  (\bibinfo{year}{2017}).
\newblock
\showeprint[arxiv]{1703.05698}
\urldef\tempurl%
\url{http://arxiv.org/abs/1703.05698}
\showURL{%
\tempurl}


\bibitem[\protect\citeauthoryear{Nguyen, Qi, Roychoudhury, and Chandra}{Nguyen
  et~al\mbox{.}}{2013}]%
        {SemFix}
\bibfield{author}{\bibinfo{person}{Hoang Duong~Thien Nguyen},
  \bibinfo{person}{Dawei Qi}, \bibinfo{person}{Abhik Roychoudhury}, {and}
  \bibinfo{person}{Satish Chandra}.} \bibinfo{year}{2013}\natexlab{}.
\newblock \showarticletitle{SemFix: Program Repair via Semantic Analysis}. In
  \bibinfo{booktitle}{\emph{Proceedings of the 2013 International Conference on
  Software Engineering}} \emph{(\bibinfo{series}{ICSE '13'})}.
  \bibinfo{publisher}{IEEE Press}, \bibinfo{address}{Piscataway, NJ, USA},
  \bibinfo{pages}{772--781}.
\newblock
\showISBNx{978-1-4673-3076-3}
\urldef\tempurl%
\url{http://dl.acm.org/citation.cfm?id=2486788.2486890}
\showURL{%
\tempurl}


\bibitem[\protect\citeauthoryear{Osera and Zdancewic}{Osera and
  Zdancewic}{2015}]%
        {OseraZ15}
\bibfield{author}{\bibinfo{person}{Peter{-}Michael Osera} {and}
  \bibinfo{person}{Steve Zdancewic}.} \bibinfo{year}{2015}\natexlab{}.
\newblock \showarticletitle{Type-and-example-directed program synthesis}. In
  \bibinfo{booktitle}{\emph{Proceedings of the 36th {ACM} {SIGPLAN} Conference
  on Programming Language Design and Implementation, Portland, OR, USA, June
  15-17, 2015}}. \bibinfo{pages}{619--630}.
\newblock
\urldef\tempurl%
\url{https://doi.org/10.1145/2737924.2738007}
\showDOI{\tempurl}


\bibitem[\protect\citeauthoryear{Peleg, Shoham, Yahav, and Yang}{Peleg
  et~al\mbox{.}}{2013}]%
        {PelegSYY13}
\bibfield{author}{\bibinfo{person}{Hila Peleg}, \bibinfo{person}{Sharon
  Shoham}, \bibinfo{person}{Eran Yahav}, {and} \bibinfo{person}{Hongseok
  Yang}.} \bibinfo{year}{2013}\natexlab{}.
\newblock \showarticletitle{Symbolic Automata for Static Specification Mining}.
  In \bibinfo{booktitle}{\emph{Static Analysis - 20th International Symposium,
  {SAS} 2013, Seattle, WA, USA, June 20-22, 2013. Proceedings}}.
  \bibinfo{pages}{63--83}.
\newblock
\urldef\tempurl%
\url{https://doi.org/10.1007/978-3-642-38856-9_6}
\showDOI{\tempurl}


\bibitem[\protect\citeauthoryear{Perelman, Gulwani, Grossman, and
  Provost}{Perelman et~al\mbox{.}}{2014}]%
        {PerelmanGGP14}
\bibfield{author}{\bibinfo{person}{Daniel Perelman}, \bibinfo{person}{Sumit
  Gulwani}, \bibinfo{person}{Dan Grossman}, {and} \bibinfo{person}{Peter
  Provost}.} \bibinfo{year}{2014}\natexlab{}.
\newblock \showarticletitle{Test-driven synthesis}. In
  \bibinfo{booktitle}{\emph{{ACM} {SIGPLAN} Conference on Programming Language
  Design and Implementation, {PLDI} '14, Edinburgh, United Kingdom - June 09 -
  11, 2014}}. \bibinfo{pages}{408--418}.
\newblock
\urldef\tempurl%
\url{https://doi.org/10.1145/2594291.2594297}
\showDOI{\tempurl}


\bibitem[\protect\citeauthoryear{Perkins, Kim, Larsen, Amarasinghe, Bachrach,
  Carbin, Pacheco, Sherwood, Sidiroglou, Sullivan, Wong, Zibin, Ernst, and
  Rinard}{Perkins et~al\mbox{.}}{2009}]%
        {ClearView}
\bibfield{author}{\bibinfo{person}{Jeff~H. Perkins}, \bibinfo{person}{Sunghun
  Kim}, \bibinfo{person}{Samuel Larsen}, \bibinfo{person}{Saman~P.
  Amarasinghe}, \bibinfo{person}{Jonathan Bachrach}, \bibinfo{person}{Michael
  Carbin}, \bibinfo{person}{Carlos Pacheco}, \bibinfo{person}{Frank Sherwood},
  \bibinfo{person}{Stelios Sidiroglou}, \bibinfo{person}{Greg Sullivan},
  \bibinfo{person}{Weng{-}Fai Wong}, \bibinfo{person}{Yoav Zibin},
  \bibinfo{person}{Michael~D. Ernst}, {and} \bibinfo{person}{Martin~C.
  Rinard}.} \bibinfo{year}{2009}\natexlab{}.
\newblock \showarticletitle{Automatically patching errors in deployed
  software}. In \bibinfo{booktitle}{\emph{Proceedings of the 22nd {ACM}
  Symposium on Operating Systems Principles 2009, {SOSP} 2009, Big Sky,
  Montana, USA, October 11-14, 2009}}. \bibinfo{pages}{87--102}.
\newblock


\bibitem[\protect\citeauthoryear{Phothilimthana, Jelvis, Shah, Totla, Chasins,
  and Bod{\'{\i}}k}{Phothilimthana et~al\mbox{.}}{2014}]%
        {PhothilimthanaJSTCB14}
\bibfield{author}{\bibinfo{person}{Phitchaya~Mangpo Phothilimthana},
  \bibinfo{person}{Tikhon Jelvis}, \bibinfo{person}{Rohin Shah},
  \bibinfo{person}{Nishant Totla}, \bibinfo{person}{Sarah Chasins}, {and}
  \bibinfo{person}{Rastislav Bod{\'{\i}}k}.} \bibinfo{year}{2014}\natexlab{}.
\newblock \showarticletitle{Chlorophyll: synthesis-aided compiler for low-power
  spatial architectures}. In \bibinfo{booktitle}{\emph{{ACM} {SIGPLAN}
  Conference on Programming Language Design and Implementation, {PLDI} '14,
  Edinburgh, United Kingdom - June 09 - 11, 2014}}. \bibinfo{pages}{396--407}.
\newblock
\urldef\tempurl%
\url{https://doi.org/10.1145/2594291.2594339}
\showDOI{\tempurl}


\bibitem[\protect\citeauthoryear{Polikarpova, Kuraj, and
  Solar{-}Lezama}{Polikarpova et~al\mbox{.}}{2016}]%
        {PolikarpovaKS16}
\bibfield{author}{\bibinfo{person}{Nadia Polikarpova}, \bibinfo{person}{Ivan
  Kuraj}, {and} \bibinfo{person}{Armando Solar{-}Lezama}.}
  \bibinfo{year}{2016}\natexlab{}.
\newblock \showarticletitle{Program synthesis from polymorphic refinement
  types}. In \bibinfo{booktitle}{\emph{Proceedings of the 37th {ACM} {SIGPLAN}
  Conference on Programming Language Design and Implementation, {PLDI} 2016,
  Santa Barbara, CA, USA, June 13-17, 2016}}. \bibinfo{pages}{522--538}.
\newblock
\urldef\tempurl%
\url{https://doi.org/10.1145/2908080.2908093}
\showDOI{\tempurl}


\bibitem[\protect\citeauthoryear{Polozov and Gulwani}{Polozov and
  Gulwani}{2015}]%
        {PolozovG15}
\bibfield{author}{\bibinfo{person}{Oleksandr Polozov} {and}
  \bibinfo{person}{Sumit Gulwani}.} \bibinfo{year}{2015}\natexlab{}.
\newblock \showarticletitle{FlashMeta: a framework for inductive program
  synthesis}. In \bibinfo{booktitle}{\emph{Proceedings of the 2015 {ACM}
  {SIGPLAN} International Conference on Object-Oriented Programming, Systems,
  Languages, and Applications, {OOPSLA} 2015, part of {SPLASH} 2015,
  Pittsburgh, PA, USA, October 25-30, 2015}}. \bibinfo{pages}{107--126}.
\newblock
\urldef\tempurl%
\url{https://doi.org/10.1145/2814270.2814310}
\showDOI{\tempurl}


\bibitem[\protect\citeauthoryear{Popa, Redfield, Zeldovich, and
  Balakrishnan}{Popa et~al\mbox{.}}{2011}]%
        {popa:2011:sosp}
\bibfield{author}{\bibinfo{person}{Raluca~Ada Popa}, \bibinfo{person}{Catherine
  M.~S. Redfield}, \bibinfo{person}{Nickolai Zeldovich}, {and}
  \bibinfo{person}{Hari Balakrishnan}.} \bibinfo{year}{2011}\natexlab{}.
\newblock \showarticletitle{CryptDB: Protecting Confidentiality with Encrypted
  Query Processing}. In \bibinfo{booktitle}{\emph{Proceedings of the
  Twenty-Third ACM Symposium on Operating Systems Principles}}
  \emph{(\bibinfo{series}{SOSP '11})}. \bibinfo{publisher}{ACM},
  \bibinfo{address}{New York, NY, USA}, \bibinfo{pages}{85--100}.
\newblock
\showISBNx{978-1-4503-0977-6}
\urldef\tempurl%
\url{https://doi.org/10.1145/2043556.2043566}
\showDOI{\tempurl}


\bibitem[\protect\citeauthoryear{P{\"{u}}schel, Moura, Singer, Xiong, Johnson,
  Padua, Veloso, and Johnson}{P{\"{u}}schel et~al\mbox{.}}{2004}]%
        {PuschelMSXJPVJ04}
\bibfield{author}{\bibinfo{person}{Markus P{\"{u}}schel},
  \bibinfo{person}{Jos{\'{e}} M.~F. Moura}, \bibinfo{person}{Bryan Singer},
  \bibinfo{person}{Jianxin Xiong}, \bibinfo{person}{Jeremy~R. Johnson},
  \bibinfo{person}{David~A. Padua}, \bibinfo{person}{Manuela~M. Veloso}, {and}
  \bibinfo{person}{Robert~W. Johnson}.} \bibinfo{year}{2004}\natexlab{}.
\newblock \showarticletitle{Spiral: {A} Generator for Platform-Adapted
  Libraries of Signal Processing Alogorithms}.
\newblock \bibinfo{journal}{\emph{{IJHPCA}}} \bibinfo{volume}{18},
  \bibinfo{number}{1} (\bibinfo{year}{2004}), \bibinfo{pages}{21--45}.
\newblock
\urldef\tempurl%
\url{https://doi.org/10.1177/1094342004041291}
\showDOI{\tempurl}


\bibitem[\protect\citeauthoryear{Ragan{-}Kelley, Barnes, Adams, Paris, Durand,
  and Amarasinghe}{Ragan{-}Kelley et~al\mbox{.}}{2013}]%
        {Ragan-KelleyBAPDA13}
\bibfield{author}{\bibinfo{person}{Jonathan Ragan{-}Kelley},
  \bibinfo{person}{Connelly Barnes}, \bibinfo{person}{Andrew Adams},
  \bibinfo{person}{Sylvain Paris}, \bibinfo{person}{Fr{\'{e}}do Durand}, {and}
  \bibinfo{person}{Saman~P. Amarasinghe}.} \bibinfo{year}{2013}\natexlab{}.
\newblock \showarticletitle{Halide: a language and compiler for optimizing
  parallelism, locality, and recomputation in image processing pipelines}. In
  \bibinfo{booktitle}{\emph{{ACM} {SIGPLAN} Conference on Programming Language
  Design and Implementation, {PLDI} '13, Seattle, WA, USA, June 16-19, 2013}}.
  \bibinfo{pages}{519--530}.
\newblock
\urldef\tempurl%
\url{https://doi.org/10.1145/2462156.2462176}
\showDOI{\tempurl}


\bibitem[\protect\citeauthoryear{Raychev, Vechev, and Krause}{Raychev
  et~al\mbox{.}}{2015}]%
        {Raychev:2015:PPP}
\bibfield{author}{\bibinfo{person}{Veselin Raychev}, \bibinfo{person}{Martin
  Vechev}, {and} \bibinfo{person}{Andreas Krause}.}
  \bibinfo{year}{2015}\natexlab{}.
\newblock \showarticletitle{Predicting Program Properties from "Big Code"}. In
  \bibinfo{booktitle}{\emph{Proceedings of the 42Nd Annual ACM SIGPLAN-SIGACT
  Symposium on Principles of Programming Languages}}
  \emph{(\bibinfo{series}{POPL '15})}. \bibinfo{publisher}{ACM},
  \bibinfo{address}{New York, NY, USA}, \bibinfo{pages}{111--124}.
\newblock
\showISBNx{978-1-4503-3300-9}
\urldef\tempurl%
\url{https://doi.org/10.1145/2676726.2677009}
\showDOI{\tempurl}


\bibitem[\protect\citeauthoryear{Re}{Re}{2017}]%
        {babble:re:2017}
\bibfield{author}{\bibinfo{person}{Chris Re}.} \bibinfo{year}{2017}\natexlab{}.
\newblock \bibinfo{booktitle}{\emph{Babble Labble}}.
\newblock \bibinfo{type}{{T}echnical {R}eport}.
  \bibinfo{institution}{Department of Computer Science, Stanford},
  \bibinfo{address}{Stanford, CA}.
\newblock
\urldef\tempurl%
\url{https://hazyresearch.github.io/snorkel/blog/babble_labble.html}
\showURL{%
\tempurl}


\bibitem[\protect\citeauthoryear{Rich and Waters}{Rich and Waters}{1988}]%
        {rich:1988:par}
\bibfield{author}{\bibinfo{person}{Charles Rich} {and}
  \bibinfo{person}{Richard~C. Waters}.} \bibinfo{year}{1988}\natexlab{}.
\newblock \showarticletitle{The Programmer's Apprentice: A Research Overview}.
\newblock \bibinfo{journal}{\emph{Computer}} \bibinfo{volume}{21},
  \bibinfo{number}{11} (\bibinfo{date}{Nov.} \bibinfo{year}{1988}),
  \bibinfo{pages}{10--25}.
\newblock
\showISSN{0018-9162}
\urldef\tempurl%
\url{https://doi.org/10.1109/2.86782}
\showDOI{\tempurl}


\bibitem[\protect\citeauthoryear{Rolim, Soares, D'Antoni, Polozov, Gulwani,
  Gheyi, Suzuki, and Hartmann}{Rolim et~al\mbox{.}}{2017}]%
        {REFAZER}
\bibfield{author}{\bibinfo{person}{Reudismam Rolim}, \bibinfo{person}{Gustavo
  Soares}, \bibinfo{person}{Loris D'Antoni}, \bibinfo{person}{Oleksandr
  Polozov}, \bibinfo{person}{Sumit Gulwani}, \bibinfo{person}{Rohit Gheyi},
  \bibinfo{person}{Ryo Suzuki}, {and} \bibinfo{person}{Bj{\"{o}}rn Hartmann}.}
  \bibinfo{year}{2017}\natexlab{}.
\newblock \showarticletitle{Learning syntactic program transformations from
  examples}. In \bibinfo{booktitle}{\emph{Proceedings of the 39th International
  Conference on Software Engineering, {ICSE} 2017, Buenos Aires, Argentina, May
  20-28, 2017}}.
\newblock


\bibitem[\protect\citeauthoryear{Sidiroglou{-}Douskos, Lahtinen, Long, and
  Rinard}{Sidiroglou{-}Douskos et~al\mbox{.}}{2015a}]%
        {CodePhage}
\bibfield{author}{\bibinfo{person}{Stelios Sidiroglou{-}Douskos},
  \bibinfo{person}{Eric Lahtinen}, \bibinfo{person}{Fan Long}, {and}
  \bibinfo{person}{Martin Rinard}.} \bibinfo{year}{2015}\natexlab{a}.
\newblock \showarticletitle{Automatic error elimination by horizontal code
  transfer across multiple applications}. In
  \bibinfo{booktitle}{\emph{Proceedings of the 36th {ACM} {SIGPLAN} Conference
  on Programming Language Design and Implementation, Portland, OR, USA, June
  15-17, 2015}}. \bibinfo{pages}{43--54}.
\newblock


\bibitem[\protect\citeauthoryear{Sidiroglou{-}Douskos, Lahtinen, Rittenhouse,
  Piselli, Long, Kim, and Rinard}{Sidiroglou{-}Douskos et~al\mbox{.}}{2015b}]%
        {DIODE}
\bibfield{author}{\bibinfo{person}{Stelios Sidiroglou{-}Douskos},
  \bibinfo{person}{Eric Lahtinen}, \bibinfo{person}{Nathan Rittenhouse},
  \bibinfo{person}{Paolo Piselli}, \bibinfo{person}{Fan Long},
  \bibinfo{person}{Deokhwan Kim}, {and} \bibinfo{person}{Martin~C. Rinard}.}
  \bibinfo{year}{2015}\natexlab{b}.
\newblock \showarticletitle{Targeted Automatic Integer Overflow Discovery Using
  Goal-Directed Conditional Branch Enforcement}. In
  \bibinfo{booktitle}{\emph{Proceedings of the Twentieth International
  Conference on Architectural Support for Programming Languages and Operating
  Systems, {ASPLOS} '15, Istanbul, Turkey, March 14-18, 2015}}.
  \bibinfo{pages}{473--486}.
\newblock


\bibitem[\protect\citeauthoryear{Sidiroglou-Douskos, Lantinen, Eden, Long, and
  Rinard}{Sidiroglou-Douskos et~al\mbox{.}}{2017}]%
        {CodeCarbonCopy}
\bibfield{author}{\bibinfo{person}{Stelios Sidiroglou-Douskos},
  \bibinfo{person}{Eric Lantinen}, \bibinfo{person}{Anthony Eden},
  \bibinfo{person}{Fan Long}, {and} \bibinfo{person}{Martin Rinard}.}
  \bibinfo{year}{2017}\natexlab{}.
\newblock \showarticletitle{CodeCarbonCopy}. In
  \bibinfo{booktitle}{\emph{Proceedings of the 2017 11th Joint Meeting on
  Foundations of Software Engineering}} \emph{(\bibinfo{series}{ESEC/FSE
  2017})}.
\newblock


\bibitem[\protect\citeauthoryear{Singh and Gulwani}{Singh and Gulwani}{2015}]%
        {SinghG15}
\bibfield{author}{\bibinfo{person}{Rishabh Singh} {and} \bibinfo{person}{Sumit
  Gulwani}.} \bibinfo{year}{2015}\natexlab{}.
\newblock \showarticletitle{Predicting a Correct Program in Programming by
  Example}. In \bibinfo{booktitle}{\emph{Computer Aided Verification - 27th
  International Conference, {CAV} 2015, San Francisco, CA, USA, July 18-24,
  2015, Proceedings, Part {I}}}. \bibinfo{pages}{398--414}.
\newblock
\urldef\tempurl%
\url{https://doi.org/10.1007/978-3-319-21690-4_23}
\showDOI{\tempurl}


\bibitem[\protect\citeauthoryear{Singh and Solar{-}Lezama}{Singh and
  Solar{-}Lezama}{2011}]%
        {SinghS11}
\bibfield{author}{\bibinfo{person}{Rishabh Singh} {and}
  \bibinfo{person}{Armando Solar{-}Lezama}.} \bibinfo{year}{2011}\natexlab{}.
\newblock \showarticletitle{Synthesizing data structure manipulations from
  storyboards}. In \bibinfo{booktitle}{\emph{SIGSOFT/FSE'11 19th {ACM}
  {SIGSOFT} Symposium on the Foundations of Software Engineering {(FSE-19)} and
  ESEC'11: 13th European Software Engineering Conference (ESEC-13), Szeged,
  Hungary, September 5-9, 2011}}. \bibinfo{pages}{289--299}.
\newblock
\urldef\tempurl%
\url{https://doi.org/10.1145/2025113.2025153}
\showDOI{\tempurl}


\bibitem[\protect\citeauthoryear{Singh and Solar{-}Lezama}{Singh and
  Solar{-}Lezama}{2012}]%
        {SinghS12}
\bibfield{author}{\bibinfo{person}{Rishabh Singh} {and}
  \bibinfo{person}{Armando Solar{-}Lezama}.} \bibinfo{year}{2012}\natexlab{}.
\newblock \showarticletitle{{SPT:} Storyboard Programming Tool}. In
  \bibinfo{booktitle}{\emph{Computer Aided Verification - 24th International
  Conference, {CAV} 2012, Berkeley, CA, USA, July 7-13, 2012 Proceedings}}.
  \bibinfo{pages}{738--743}.
\newblock
\urldef\tempurl%
\url{https://doi.org/10.1007/978-3-642-31424-7_58}
\showDOI{\tempurl}


\bibitem[\protect\citeauthoryear{Solar{-}Lezama, Tancau, Bod{\'{\i}}k, Seshia,
  and Saraswat}{Solar{-}Lezama et~al\mbox{.}}{2006}]%
        {Solar-LezamaTBSS06}
\bibfield{author}{\bibinfo{person}{Armando Solar{-}Lezama},
  \bibinfo{person}{Liviu Tancau}, \bibinfo{person}{Rastislav Bod{\'{\i}}k},
  \bibinfo{person}{Sanjit~A. Seshia}, {and} \bibinfo{person}{Vijay~A.
  Saraswat}.} \bibinfo{year}{2006}\natexlab{}.
\newblock \showarticletitle{Combinatorial sketching for finite programs}. In
  \bibinfo{booktitle}{\emph{Proceedings of the 12th International Conference on
  Architectural Support for Programming Languages and Operating Systems,
  {ASPLOS} 2006, San Jose, CA, USA, October 21-25, 2006}}.
  \bibinfo{pages}{404--415}.
\newblock
\urldef\tempurl%
\url{https://doi.org/10.1145/1168857.1168907}
\showDOI{\tempurl}


\bibitem[\protect\citeauthoryear{Son, McKinley, and Shmatikov}{Son
  et~al\mbox{.}}{2013}]%
        {FixMeUp}
\bibfield{author}{\bibinfo{person}{Sooel Son}, \bibinfo{person}{Kathryn~S
  McKinley}, {and} \bibinfo{person}{Vitaly Shmatikov}.}
  \bibinfo{year}{2013}\natexlab{}.
\newblock \showarticletitle{Fix Me Up: Repairing Access-Control Bugs in Web
  Applications.}. In \bibinfo{booktitle}{\emph{NDSS}}.
\newblock


\bibitem[\protect\citeauthoryear{Srivastava, Gulwani, and Foster}{Srivastava
  et~al\mbox{.}}{2010}]%
        {SrivastavaGF10}
\bibfield{author}{\bibinfo{person}{Saurabh Srivastava}, \bibinfo{person}{Sumit
  Gulwani}, {and} \bibinfo{person}{Jeffrey~S. Foster}.}
  \bibinfo{year}{2010}\natexlab{}.
\newblock \showarticletitle{From program verification to program synthesis}. In
  \bibinfo{booktitle}{\emph{Proceedings of the 37th {ACM} {SIGPLAN-SIGACT}
  Symposium on Principles of Programming Languages, {POPL} 2010, Madrid, Spain,
  January 17-23, 2010}}. \bibinfo{pages}{313--326}.
\newblock
\urldef\tempurl%
\url{https://doi.org/10.1145/1706299.1706337}
\showDOI{\tempurl}


\bibitem[\protect\citeauthoryear{Udupa, Raghavan, Deshmukh, Mador{-}Haim,
  Martin, and Alur}{Udupa et~al\mbox{.}}{2013}]%
        {UdupaRDMMA13}
\bibfield{author}{\bibinfo{person}{Abhishek Udupa}, \bibinfo{person}{Arun
  Raghavan}, \bibinfo{person}{Jyotirmoy~V. Deshmukh}, \bibinfo{person}{Sela
  Mador{-}Haim}, \bibinfo{person}{Milo M.~K. Martin}, {and}
  \bibinfo{person}{Rajeev Alur}.} \bibinfo{year}{2013}\natexlab{}.
\newblock \showarticletitle{{TRANSIT:} specifying protocols with concolic
  snippets}. In \bibinfo{booktitle}{\emph{{ACM} {SIGPLAN} Conference on
  Programming Language Design and Implementation, {PLDI} '13, Seattle, WA, USA,
  June 16-19, 2013}}. \bibinfo{pages}{287--296}.
\newblock
\urldef\tempurl%
\url{https://doi.org/10.1145/2462156.2462174}
\showDOI{\tempurl}


\bibitem[\protect\citeauthoryear{Vechev and Yahav}{Vechev and Yahav}{2016}]%
        {vechev:2016:pbigcode}
\bibfield{author}{\bibinfo{person}{Martin Vechev} {and} \bibinfo{person}{Eran
  Yahav}.} \bibinfo{year}{2016}\natexlab{}.
\newblock \showarticletitle{Programming with "Big Code"}.
\newblock \bibinfo{journal}{\emph{Found. Trends Program. Lang.}}
  \bibinfo{volume}{3}, \bibinfo{number}{4} (\bibinfo{date}{Dec.}
  \bibinfo{year}{2016}), \bibinfo{pages}{231--284}.
\newblock
\showISSN{2325-1107}
\urldef\tempurl%
\url{https://doi.org/10.1561/2500000028}
\showDOI{\tempurl}


\bibitem[\protect\citeauthoryear{Vechev and Yahav}{Vechev and Yahav}{2008}]%
        {VechevY08}
\bibfield{author}{\bibinfo{person}{Martin~T. Vechev} {and}
  \bibinfo{person}{Eran Yahav}.} \bibinfo{year}{2008}\natexlab{}.
\newblock \showarticletitle{Deriving linearizable fine-grained concurrent
  objects}. In \bibinfo{booktitle}{\emph{Proceedings of the {ACM} {SIGPLAN}
  2008 Conference on Programming Language Design and Implementation, Tucson,
  AZ, USA, June 7-13, 2008}}. \bibinfo{pages}{125--135}.
\newblock
\urldef\tempurl%
\url{https://doi.org/10.1145/1375581.1375598}
\showDOI{\tempurl}


\bibitem[\protect\citeauthoryear{Weimer, Nguyen, {Le Goues}, and
  Forrest}{Weimer et~al\mbox{.}}{2009}]%
        {WeimerNGF09}
\bibfield{author}{\bibinfo{person}{Westley Weimer}, \bibinfo{person}{ThanhVu
  Nguyen}, \bibinfo{person}{Claire {Le Goues}}, {and}
  \bibinfo{person}{Stephanie Forrest}.} \bibinfo{year}{2009}\natexlab{}.
\newblock \showarticletitle{Automatically finding patches using genetic
  programming}. In \bibinfo{booktitle}{\emph{31st International Conference on
  Software Engineering, {ICSE} 2009, May 16-24, 2009, Vancouver, Canada,
  Proceedings}}. \bibinfo{pages}{364--374}.
\newblock
\urldef\tempurl%
\url{https://doi.org/10.1109/ICSE.2009.5070536}
\showDOI{\tempurl}


\bibitem[\protect\citeauthoryear{Whaley and Dongarra}{Whaley and
  Dongarra}{1998}]%
        {WhaleyD98}
\bibfield{author}{\bibinfo{person}{R.~Clinton Whaley} {and}
  \bibinfo{person}{Jack~J. Dongarra}.} \bibinfo{year}{1998}\natexlab{}.
\newblock \showarticletitle{Automatically Tuned Linear Algebra Software}. In
  \bibinfo{booktitle}{\emph{Proceedings of the {ACM/IEEE} Conference on
  Supercomputing, {SC} 1998, November 7-13, 1998, Orlando, FL, {USA}}}.
  \bibinfo{pages}{38}.
\newblock
\urldef\tempurl%
\url{https://doi.org/10.1109/SC.1998.10004}
\showDOI{\tempurl}


\bibitem[\protect\citeauthoryear{Xiong, Wang, Yan, Zhang, Han, Huang, and
  Zhang}{Xiong et~al\mbox{.}}{2017}]%
        {ACS}
\bibfield{author}{\bibinfo{person}{Yingfei Xiong}, \bibinfo{person}{Jie Wang},
  \bibinfo{person}{Runfa Yan}, \bibinfo{person}{Jiachen Zhang},
  \bibinfo{person}{Shi Han}, \bibinfo{person}{Gang Huang}, {and}
  \bibinfo{person}{Lu Zhang}.} \bibinfo{year}{2017}\natexlab{}.
\newblock \showarticletitle{Precise Condition Synthesis for Program Repair}. In
  \bibinfo{booktitle}{\emph{Proceedings of the 39th International Conference on
  Software Engineering}} \emph{(\bibinfo{series}{ICSE '17})}.
\newblock
\showISBNx{978-1-5386-3868-2}


\bibitem[\protect\citeauthoryear{Xu, Liu, and Song}{Xu et~al\mbox{.}}{2017}]%
        {xu:2017:natural}
\bibfield{author}{\bibinfo{person}{Xiaojun Xu}, \bibinfo{person}{Chang Liu},
  {and} \bibinfo{person}{Dawn Song}.} \bibinfo{year}{2017}\natexlab{}.
\newblock \showarticletitle{SQLNet: Generating Structured Queries From Natural
  Language Without Reinforcement Learning}.
\newblock \bibinfo{journal}{\emph{CoRR}}  \bibinfo{volume}{abs/1711.04436}
  (\bibinfo{year}{2017}).
\newblock
\showeprint[arxiv]{1711.04436}
\urldef\tempurl%
\url{http://arxiv.org/abs/1711.04436}
\showURL{%
\tempurl}


\bibitem[\protect\citeauthoryear{Yadid and Yahav}{Yadid and Yahav}{2016}]%
        {Yadid:2016:ECP}
\bibfield{author}{\bibinfo{person}{Shir Yadid} {and} \bibinfo{person}{Eran
  Yahav}.} \bibinfo{year}{2016}\natexlab{}.
\newblock \showarticletitle{Extracting Code from Programming Tutorial Videos}.
  In \bibinfo{booktitle}{\emph{Proceedings of the 2016 ACM International
  Symposium on New Ideas, New Paradigms, and Reflections on Programming and
  Software}} \emph{(\bibinfo{series}{Onward! 2016})}. \bibinfo{publisher}{ACM},
  \bibinfo{address}{New York, NY, USA}, \bibinfo{pages}{98--111}.
\newblock
\showISBNx{978-1-4503-4076-2}
\urldef\tempurl%
\url{https://doi.org/10.1145/2986012.2986021}
\showDOI{\tempurl}


\bibitem[\protect\citeauthoryear{Yaghmazadeh, Wang, Dillig, and
  Dillig}{Yaghmazadeh et~al\mbox{.}}{2017}]%
        {Yaghmazadeh0DD17}
\bibfield{author}{\bibinfo{person}{Navid Yaghmazadeh}, \bibinfo{person}{Yuepeng
  Wang}, \bibinfo{person}{Isil Dillig}, {and} \bibinfo{person}{Thomas Dillig}.}
  \bibinfo{year}{2017}\natexlab{}.
\newblock \showarticletitle{SQLizer: query synthesis from natural language}.
\newblock \bibinfo{journal}{\emph{{PACMPL}}} \bibinfo{volume}{1},
  \bibinfo{number}{{OOPSLA}} (\bibinfo{year}{2017}),
  \bibinfo{pages}{63:1--63:26}.
\newblock
\urldef\tempurl%
\url{https://doi.org/10.1145/3133887}
\showDOI{\tempurl}


\bibitem[\protect\citeauthoryear{Zheng, Dave, Beekman, Popa, Gonzalez, and
  Stoica}{Zheng et~al\mbox{.}}{2017}]%
        {zheng:2017:opaque}
\bibfield{author}{\bibinfo{person}{Wenting Zheng}, \bibinfo{person}{Ankur
  Dave}, \bibinfo{person}{Jethro~G. Beekman}, \bibinfo{person}{Raluca~Ada
  Popa}, \bibinfo{person}{Joseph~E. Gonzalez}, {and} \bibinfo{person}{Ion
  Stoica}.} \bibinfo{year}{2017}\natexlab{}.
\newblock \showarticletitle{Opaque: An Oblivious and Encrypted Distributed
  Analytics Platform}. In \bibinfo{booktitle}{\emph{14th {USENIX} Symposium on
  Networked Systems Design and Implementation, {NSDI} 2017, Boston, MA, USA,
  March 27-29, 2017}}. \bibinfo{pages}{283--298}.
\newblock
\urldef\tempurl%
\url{https://www.usenix.org/conference/nsdi17/technical-sessions/presentation/zheng}
\showURL{%
\tempurl}


\end{thebibliography}

\end{document}